\pdfoutput=1

\documentclass[11pt]{article}

\usepackage{emnlp2022}

\usepackage{times}
\usepackage{latexsym}
\usepackage{amsmath}
\usepackage{amsfonts}
\usepackage{amssymb}
\usepackage{amsthm}
\usepackage{multicol}
\usepackage{booktabs}
\usepackage{flushend}
\usepackage{inconsolata}

\usepackage[T5,T1]{fontenc}
\usepackage[utf8]{inputenc}

\usepackage{microtype}
\usepackage{bm}

\DeclareTextSymbolDefault{\ohorn}{T5}
\DeclareTextSymbolDefault{\uhorn}{T5}

\usepackage{caption}
\usepackage{subcaption}
\usepackage{tikz}
\usepackage[normalem]{ulem}

\usepackage{cleveref}
\crefname{section}{\S}{\S\S}
\Crefname{section}{\S}{\S\S}
\crefname{table}{Tab.}{}
\crefname{figure}{Fig.}{Figs.}
\crefname{equation}{Eq.}{Eqs.}
\crefname{algorithm}{Algorithm}{}
\crefname{line}{Line}{}
\crefname{appendix}{App.}{}
\crefformat{section}{\S#2#1#3}
\crefname{thm}{Theorem}{}
\crefname{cor}{Corollary}{}
\crefname{prop}{Proposition}{}
\crefname{def}{Definition}{}

\newtheorem{principle}{}

\newcommand{\defn}[1]{\textbf{#1}}

\newcommand{\calV}{\mathcal{V}}
\newcommand{\calA}{\mathcal{A}}
\newcommand{\calR}{\mathcal{R}}
\newcommand{\vinfo}{$\calV$-information}
\newcommand{\binfo}{$\bm{\calV}$\textbf{-information}}
\newcommand{\MI}{\mathrm{I}}
\newcommand{\vMI}{\MI_{\calV}}
\newcommand{\ent}{\mathrm{H}}
\newcommand{\vent}{\ent_{\calV}}

\newcommand{\vcoefficient}{$\calV$-coefficient}
\newcommand{\bcoefficient}{$\bm{\calV}$-coefficient}
\newcommand{\vcoeff}{\mathrm{C}_{\calV}}

\newcommand{\dataset}{\mathcal{D}}
\newcommand{\trainset}{\dataset_{\mathrm{train}}}
\newcommand{\devset}{\dataset_{\mathrm{dev}}}
\newcommand{\testset}{\dataset_{\mathrm{test}}}
\newcommand{\loss}{\mathcal{L}}
\newcommand{\btheta}{{\boldsymbol \theta}}
\newcommand{\qtheta}{q_{\btheta}}
\newcommand{\xent}{\ent_{\btheta}}

\newcommand{\R}{\mathbb{R}}
\newcommand{\bw}{\mathbf{w}}
\newcommand{\nbw}{\mathrm{w}}
\newcommand{\bs}{\mathbf{s}}
\newcommand{\br}{\mathbf{r}}
\newcommand{\bp}{\mathbf{p}}
\newcommand{\ba}{\mathbf{a}}
\newcommand{\bK}{\mathbf{K}}
\newcommand{\bB}{\mathbf{B}}
\newcommand{\bQ}{\mathbf{Q}}
\newcommand{\bW}{\mathbf{W}}
\newcommand{\balpha}{{\boldsymbol \alpha}}
\newcommand{\layer}{\ell}
\newcommand{\MLP}{\mathrm{MLP}}

\newcommand{\bR}{\mathbf{R}}
\newcommand{\bA}{\mathbf{A}}
\newcommand{\bS}{\mathbf{S}}
\newcommand{\dr}{\mathrm{d}\br}

\newcommand{\defeq}[0]{\mathrel{\stackrel{\textnormal{\tiny def}}{=}}}

\title{The Architectural Bottleneck Principle}

\usepackage{emoji}
\newcommand{\ucambridge}{\emoji[emoji]{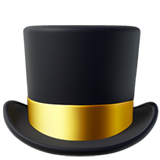}}
\newcommand{\ethz}{\emoji[emoji]{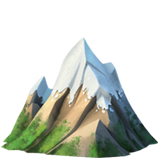}}

\author{
 Tiago Pimentel\thanks{~~Equal contribution.}$^{\,\,\,,\ucambridge}$~\;~ Josef Valvoda$^{*,\ucambridge}$~\;~ Niklas Stoehr$^{\ethz}$~\;~ Ryan Cotterell$^{\ethz}$ \\
 $^{\ucambridge}$University of Cambridge~\;~  $^{\ethz}$ETH Zürich\\
 \texttt{\{\href{mailto:tp472@cam.ac.uk}{tp472}, \href{mailto:jv406@cam.ac.uk}{jv406}\}@cam.ac.uk} \\
 \texttt{\{\href{mailto:niklas.stoehr@inf.ethz.ch}{niklas.stoehr}, \href{mailto:ryan.cotterell@inf.ethz.ch}{ryan.cotterell}\}@inf.ethz.ch}
}

\begin{document}
\maketitle
\begin{abstract}
In this paper, we seek to measure how much information a component in a neural network could extract from the representations fed into it.
Our work stands in contrast to prior probing work, most of which investigates how much information a model's representations \emph{contain}.
This shift in perspective leads us to propose a new principle for probing, the \defn{architectural bottleneck principle}:
In order to estimate how much information a given component could extract, a probe should look exactly like the component.
Relying on this principle, we estimate how much syntactic information is available to transformers through our attentional probe, a probe that exactly resembles a transformer's self-attention head.
Experimentally, we find that, in three models (BERT, ALBERT, and RoBERTa), a sentence's syntax tree is mostly extractable by our probe,
suggesting these models have access to syntactic information while composing their contextual representations. 
Whether this information is actually used by these models, however, remains an open question.\looseness=-1

\vspace{0.5em}
\hspace{0em}\includegraphics[width=1.25em,height=1.25em]{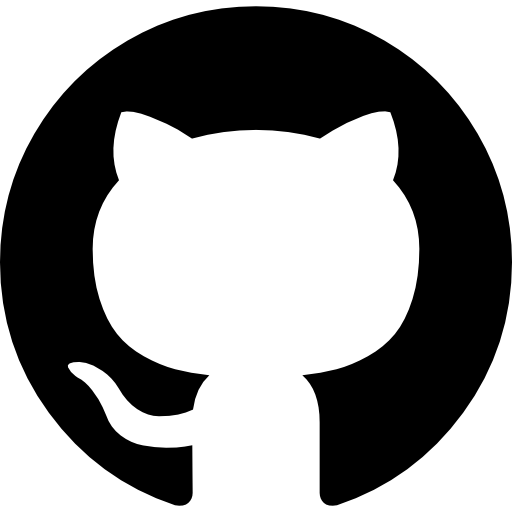}\hspace{.75em}\parbox{\dimexpr\linewidth-7\fboxsep-2\fboxrule}{\url{https://github.com/rycolab/attentional-probe}}
\vspace{-0.5em}
\end{abstract}

\section{Introduction}
The surprising performance of pretrained language models on diverse natural language processing tasks has sparked interest in their analysis. 
Probing is one of the most prevalent methods employed to engage in such an analysis.
In a typical probing study \citep[\emph{inter alia}]{alain2016understanding,belinkov-etal-2017-neural,adi2016fine}, the weights of the model under consideration are first frozen.
A probe is then trained on top of the model's contextual representations
in an attempt to predict one of the input sentence's properties, e.g., 
its syntactic parse.
Unfortunately, best practices on how to design such probes remain contested.\looseness=-1

On one side of the debate, some argue for simplicity, suggesting that simple probes are to be preferred so that we can distinguish probing from simply learning an NLP task
\citep{hewitt-liang-2019-designing}. 
On the other side of the debate, some argue we need complex probes in order to extract all relevant information from the representations \citep{saphra-lopez-2019-understanding,pimentel-etal-2020-information}.
Bridging the gap, some have also called for a compromise, advocating that all probes on the complexity--accuracy Pareto curve should be considered \citep{pimentel-etal-2020-pareto}.\looseness=-1
In this paper, we propose the \defn{architectural bottleneck principle} (ABP) as a guideline for constructing useful probes.
Under the ABP, a probe's architecture should mirror a component of the model being probed.
Previous work has mostly focused on how much information is contained in a set of representations.
However, if we care about whether the information is in fact used by the model, we should instead ask how much information the model in question could use.\footnote{Explicitly, we use the bigram \textit{could extract} to refer to the total amount of information a component is able to extract from the representations fed into it; this upper-bounds the information that component actually uses.\looseness=-1}
Under this perspective, the probed model's architecture acts as a natural bottleneck to how much information the model could use---and should thus also act as a constraint when probing.\footnote{For related work investigating whether a model uses some information, see \citet{ravfogel-etal-2021-counterfactual} and \citet{lasri-etal-2022-probing}.}\looseness=-1

As a concrete example, we posit that a transformer's attention head serves as a bottleneck to its use of syntactic information, as these are the only components in a transformer with access to multiple tokens at once.
Following the ABP, we thus propose the \defn{attentional probe}, which looks exactly like an attention head.
This probe allows us to answer one specific question: How much syntactic information could a transformer use while computing its attention weights?\looseness=-1

Our results reveal that most---albeit not all---syntactic information is extractable with this simple attention head architecture: 
While we estimate English sentences to contain on average $31.2$ bits of information about their syntactic tree structure, the attentional probe can extract up to $28.0$ bits.
Furthermore, while these results hold for three popular transformer-based language models (BERT, ALBERT and RoBERTa), they do not for a similar but untrained model.
This suggests that training a model shapes
its representations to encode syntactic information.
We find this trend holds across four typologically diverse languages (Basque, English, Tamil, and Turkish).
In contrast, when we keep BERT's pretrained parameters frozen and analyze the weights of its pretrained attention heads, we observe that they do not seem to encode syntax under our operationalisation.
Ergo, while we know these models could use syntactic information to compute attention weights, whether they actually do remains an open question.\looseness=-1

\section{A Taxonomy of Probing Principles}
There are many competing approaches for how to design an effective probe \cite{belinkov_survey}. 
We taxonomise them into principles here.\looseness-1

\begin{principle}
\defn{The Linearity Principle}
\citep{alain2016understanding}.
A neural network's purpose is to make information linearly separable for its final layer.
Thus, probes should be linear models.\looseness=-1
\end{principle}

\noindent 
Focusing on how much information a model could use in its final layer,\footnote{We assume throughout this paper that a model's final layer is a linear projection coupled with a softmax nonlinearity.} \citet{alain2016understanding} propose what we term the linearity principle; many subsequent studies then adopted it in designing their probes \citep{shi-etal-2016-string,ettinger-etal-2016-probing,bisazza-tump-2018-lazy,liu-etal-2019-linguistic}.
Other researchers, however, argued that a model's non-final layers do not necessarily encode information linearly \citep{conneau-etal-2018-cram,pimentel-etal-2020-information}.
They then suggested that a probe should measure the \emph{total} amount of information present in a model's representations---\emph{independent} of whether it is actually used by the model.
This led to a second principle, which we outline below.\looseness=-1

\begin{principle}
\defn{The Maximum Information Principle}
\citep{pimentel-etal-2020-information}.
A probe's goal is to estimate how much information is encoded in a set of representations. Thus, probes should be as complex as necessary to extract all relevant information.\looseness=-1
\end{principle}

\noindent 
Following this principle, some authors have found, unsurprisingly, that non-linear probes estimate larger amounts of information to be encoded in a representation than linear ones \citep{qian-etal-2016-investigating,belinkov-etal-2017-neural,white-etal-2021-non}.
\citet{pimentel-etal-2020-information}, however, argued that all contextual representations, e.g., the ones produced by BERT, encode as much information about a target attribute as the original sentence.
It follows that probing only makes sense with some constraint on probe complexity. 
Taking complexity into account suggests another natural principle for probe design.

\begin{principle}
\defn{The Easy-extraction Principle}
\citep{hewitt-liang-2019-designing}.
The goal of probing is to reveal how easy it is to extract the information encoded in the representations. 
Thus, probes should be as simple as possible without sacrificing performance.
\end{principle}

\noindent 
The idea of preferring simple probes goes by many names in the literature.
Some authors discuss the complexity of probing architectures \citep{hewitt-liang-2019-designing,voita-titov-2020-information,pimentel-etal-2020-pareto,cao-etal-2021-low}, while others discuss the amount of data required to train the probe \citep{pimentel-cotterell-2021-bayesian}.
None of the work above, however, discusses how the model actually uses the information about the target attribute \citep{elazar-etal-2021-amnesic,lasri-etal-2022-probing}.
If we are interested in whether information can be used by the model, we need a new principle.
In this work, we argue that a model's architecture should factor into the probe's design, because the model's architecture constrains the amount of information the model can use.
This leads us to propose the following principle.\looseness=-1

\begin{principle}
\defn{The Architectural Bottleneck Principle}
(ABP).
A probe should measure how much information a component of a model could use.
Thus, a probe's architecture should mirror that component.\looseness=-1
\end{principle}

\noindent
We believe the ABP naturally connects the first three principles.
Importantly, the ABP generalises the linearity principle: If a model employs a linear projection coupled with a softmax in its final layer, and our probe mirrors that layer, as linear probes do, then the ABP will be equivalent to the linearity principle.
Furthermore, the ABP also relates to the maximum information principle: If we probe a component that is expressive enough, it should be able to extract all relevant information from a set of representations.
Finally, the ABP also implicitly controls for ease of extraction by restricting the capacity of probes.

\section{Probing with Information Theory}
In this paper, we take the position that the goal of probing is to determine how much information one can extract from the representations being probed.
Following \citet{pimentel-etal-2020-information}, we now operationalise this value formally using information theory, which offers us a clean framework to quantify information.
Specifically, the measure we are interested in is a \vinfo{} \citep{xu2020theory}.\footnote{We operationalise our measure of interest as a \vinfo{} here, since \emph{information} is usually the term used to describe probing questions. Our principle, however, can be easily extended to other measures. For instance, we can define $\calV$-UAS as the supremum unlabelled attachment score (UAS) achievable by an architecture.\looseness=-1}\looseness=-1

\subsection{Mutual Information}
\citet{shannon1948mathematical} famously quantified the amount of information that a random variable ($\bR$) contains about another ($\bA$) as their mutual information\looseness=-1
\begin{equation} \label{eq:mutual_info}
\MI(\bR; \bA) \defeq \ent(\bA) - \ent(\bA \mid \bR)
\end{equation}
where $\ent(\bA)$ and $\ent(\bA \mid \bR)$ are, respectively, the entropy of $\bA$ and the conditional entropy of $\bA$ given $\bR$.
Given that $\bR$ is a continuous-valued representation with values $\br \in \calR$, and $\bA$ is a discrete-valued attribute with values $\ba \in \calA$,
these quantities are defined formally as
\begin{align}
&\ent(\bA) \defeq \sum_{\ba \in \calA} p(\ba) \log \frac{1}{p(\ba)} \\
&\ent(\bA \mid \bR) \defeq \int\limits_{\calR} \sum_{\ba \in \calA} p(\br, \ba) \log \frac{1}{p(\ba \mid \br)} \dr
\end{align}
The maximum information principle seeks to estimate \Cref{eq:mutual_info}.
Notably, the relationship between $\bR$ and $\bA$, represented by distribution $p(\ba \mid \br)$, may be arbitrarily complex, and this distributions' computational complexity has no direct effect on the conditional entropy's value $\ent(\bA \mid \bR)$.\looseness=-1

\subsection{\binfo{}} 
\label{sec:vinfo}

Under the architectural bottleneck principle, we are interested in how much information we can extract from $\bR$
about $\bA$, when constrained to only using extraction functions in a set $\calV$, the set of functions a model's component can represent.
The \vinfo{} \citep{xu2020theory}, a generalisation of \citeposs{shannon1948mathematical} mutual information, naturally operationalises this value as\looseness=-1
\begin{equation} \label{eq:vinfo}
\vMI(\bR \rightarrow \bA) \defeq \vent(\bA) - \vent(\bA \mid \bR)
\end{equation}
where the conditional $\calV$-entropy is defined as
\begin{align} 
&\vent(\bA \mid \bR) \defeq \inf_{q \in \calV}\, \int\limits_{\calR}\, \sum_{\ba \in \calA} p(\br, \ba) \log \frac{1}{q(\ba \!\mid\! \br)} \dr \label{eq:condition_vent}
\end{align}
The unconditional $\calV$-entropy is defined similarly.\looseness=-1
In words, the \vinfo{} computes the maximum information that can be extracted by a model with an architecture $\calV$.
Notably, if $\calV$ is sufficiently expressive, i.e., if $p(\ba \mid \br) \in \calV$, the \vinfo{} will be equivalent to the traditional mutual information.
Further, \vinfo{} is bounded above by the mutual information, which leads to a new value termed here the \defn{\bcoefficient{}}\looseness=-1
\begin{equation} \label{eq:vcoeff}
\vcoeff(\bA \mid \bR) \defeq \frac{\vMI(\bR \rightarrow \bA)}{\MI(\bR; \bA)}
\end{equation}
In short, the \vcoefficient{} computes the percentage of information we can extract from a random variable when restricted to variational family $\calV$.\looseness=-1

\definecolor{mypurple}{HTML}{8a2be2}
\definecolor{myyellow}{HTML}{ffa500}
\definecolor{myturquoise}{HTML}{20b2aa}

\definecolor{mygreen}{HTML}{34cd34}
\definecolor{myblue}{HTML}{4c72b0}
\definecolor{myred}{HTML}{f93537}
\definecolor{mygray}{HTML}{808080}

\newcommand{\newcolor}[2]{\textbf{\textcolor{#1}{#2}}}

\begin{figure*}[t]
\centering
\begin{subfigure}[b]{0.5\textwidth}
    \includegraphics[width=\columnwidth]{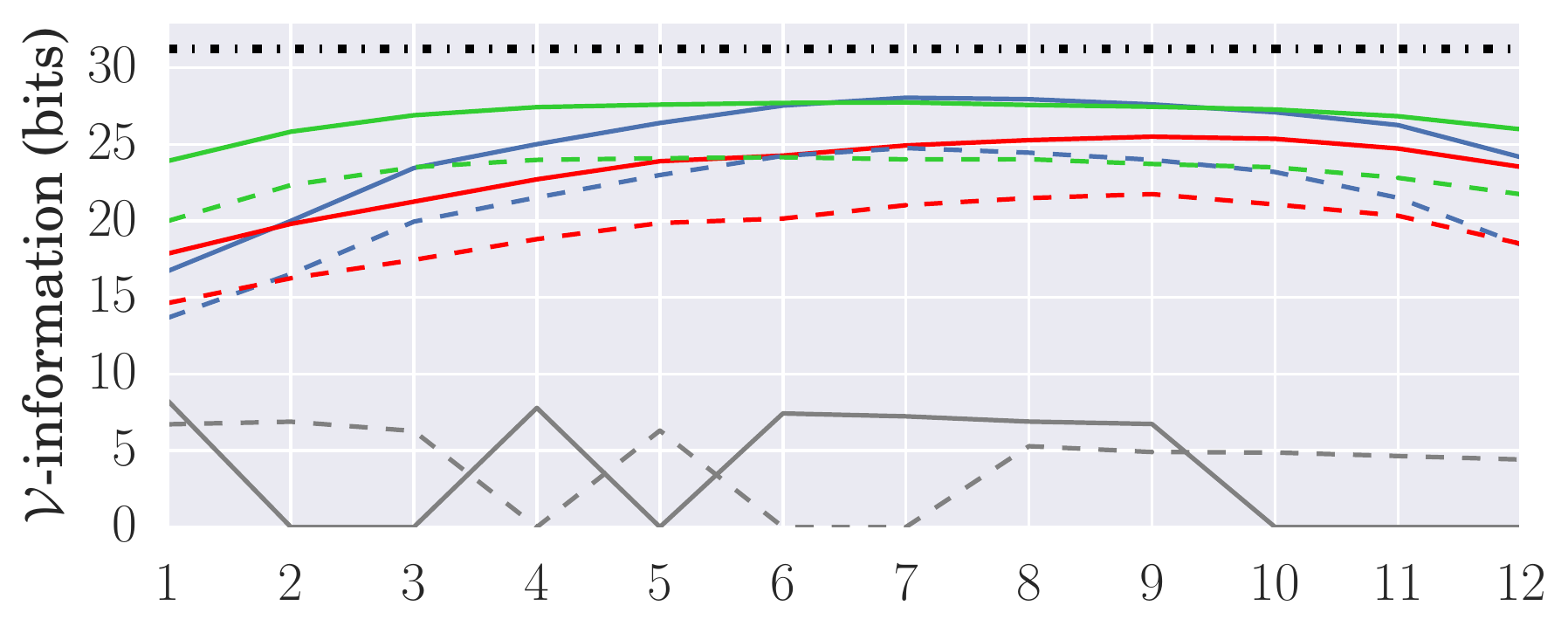}
\end{subfigure}%
~
\begin{subfigure}[b]{0.5\textwidth}
    \includegraphics[width=\columnwidth]{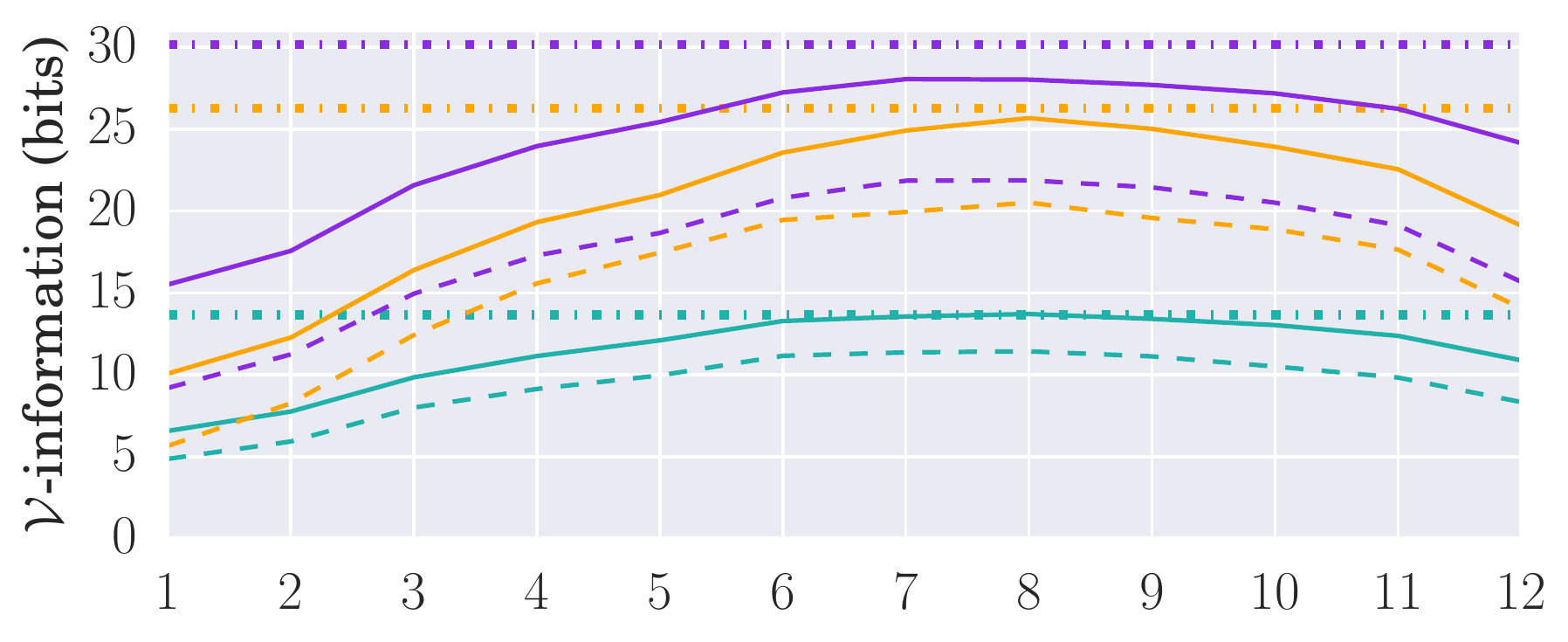}
\end{subfigure}
\caption{\vinfo{} per layer of our probes evaluated on: (left) English using \newcolor{myblue}{BERT}, \newcolor{myred}{RoBERTa}, \newcolor{mygreen}{ALBERT} and \newcolor{mygray}{untrained} representations; (right) \newcolor{mypurple}{Basque}, \newcolor{myturquoise}{Turkish}, and \newcolor{myyellow}{Tamil} using BERT representations.
Line patterns represent:
\tikz[baseline=-0.5ex]\draw [line width=.5mm,dash dot] (0,0) -- (.5,0); Mutual information;
\tikz[baseline=-0.5ex]\draw [line width=.3mm,thick] (0,0) -- (.4,0); 
Attentional \vinfo{};
\tikz[baseline=-0.5ex]\draw [line width=.3mm,densely dashed] (0,0) -- (.5,0); Structural \vinfo{}.\looseness=-1
}
\label{fig:vinfo}
\end{figure*}

\section{An Attentional Probe}

In our experiments, we will focus on a transformer's attention mechanism.
Concretely, many researchers \citep[e.g.,][]{vig-belinkov-2019-analyzing,htut_etal_2019_attention,manning2020emergent} have asserted that syntactic information is used by transformers when computing their attention weights \citep[albeit not uncontroversially; for a review, see][]{rogers_syntax}.
Further, attention heads are the only components in a transformer which have access to multiple words at the same time.
Thus, exploring the ABD in the context of attention heads is a natural starting point.
Specifically, following the ABP, we will investigate how much information a transformer's attention head could extract from the representations fed into it.

Given an input sentence $\bs$, a transformer \citep{vaswani2017attention} will generate a set of representations $\br \in \calR\!\defeq\!\R^{|\bs| \times d_1}$ at layer $\layer$.
An attention head then uses these representations to compute the attention weights
\begin{equation} \label{eq:attention_head}
    \alpha_{ij} = (\bK \br_i)^\intercal\, \bQ \br_j, 
    \quad 
    \nbw_{ij} = \frac{e^{\alpha_{ij}}}{\sum\limits_{1 \le j' \le |\bs|} e^{\alpha_{ij'}}}
\end{equation}
where $i$ and $j$ index word positions in a sentence $\bs$, $\bK, \bQ \in \R^{d_2 \times d_1}$
are the key and query matrices, and $\balpha, \bw \in \R^{|\bs| \times |\bs|}$
are, respectively, the self-attention logits and attention weights.\looseness=-1
We now consider an attentional probe parameterised using the head in \Cref{eq:attention_head}, but with randomly initialised $\bK$ and $\bQ$ matrices.
Our goal is to train this probe, as we explain towards the end of this section.
We use the attention weights, defined in \cref{eq:attention_head}, to compute the probability of a specific directed spanning tree $\ba$, which encodes the syntactic dependencies
\looseness=-1
\begin{align}
    \label{eq:tree_prob}
    \qtheta(\ba \mid \br) &= \frac{\prod_{(i,j) \in\,\ba} \nbw_{ij}}{\sum_{\ba' \in \calA_{|\bs|}} \prod_{(i,j) \in\,\ba'} \nbw_{ij}}
\end{align}
where tree $\ba$ is represented as a set of pairs $(i,j)$ which index an edge in it, $\calA_{|\bs|}$ represents the set of all directed spanning trees with a specific number of nodes $|\bs|$, and $\btheta = [\bK; \bQ] \in \mathbb{R}^{d_2 \times (2d_1)}$ represents the probe's parameters.
We can easily compute the numerator in this equation for a specific tree.
The normalising factor in the denominator is more complex, as it requires looping through a prohibitively large sum.
Luckily, we can efficiently compute it with \citeposs{koo-etal-2007-structured} variation of the matrix--tree theorem (MTT)
 for root-constrained directed spanning trees \citep{tutte1984graph,zmigrod-etal-2020-please}.\footnote{
Importantly, \citeposs{koo-etal-2007-structured} method requires a set of weights between each word and a sentence's root.
To handle this, we feed an extra root representation $\br_{0}$, initialised as a vector with all zeros, to our attentional probe, making our attention weights actually have shape $\bw \in \R^{|\bs|\!+\!1 \times |\bs|\!+\!1}$.
Explicitly, adding a root to an undirected dependency tree is equivalent to making it directed.
We then use \citeposs{zmigrod-etal-2021-efficient-computation} implementation of the matrix--tree theorem.}\looseness=-1

\paragraph{The Variational Family.} 
The attentional probe architecture is defined by \Cref{eq:attention_head,eq:tree_prob}.
We now define the equivalent variational family\looseness=-1%
\begin{equation}
    \calV = \left\{\qtheta(\ba \mid \br) \mid \bK, \bQ \in \R^{d_2 \times d_1} \right\}
\end{equation}
This variational family includes the set of all distributions computable by an attention head architecture.
In practice, however, we cannot compute the infimum over the set $\calV$ as required in \Cref{eq:condition_vent}.
As an approximation, we train our attentional probe to minimise a cross-entropy loss, which gives us an estimate of the $\calV$-entropy.
We expand on this point in \Cref{sec:vinfo_and_probing}.\footnote{We note that \citet{hewitt-etal-2021-conditional} first noted the equivalence between estimating a \vinfo{} and probing.}\looseness=-1

\section{Experiments}

\vspace{-1pt}
\paragraph{Data.}
We use the universal dependencies' (UD) treebanks \citep{ud-2.6}.
Specifically, we analyse results in four typologically diverse languages: Basque, English, Tamil, and Turkish.
Furthermore, we focus our analysis on unlabelled dependency trees.
We note that UD uses a particular syntax formalism, which could impact our results \citep{kuznetsov-gurevych-2020-matter}.

\vspace{-1pt}
\paragraph{Models.}
Empirically, we study multilingual BERT in all four languages under consideration \citep{devlin-etal-2019-bert} as well as RoBERTa and ALBERT \citep{liu2019roberta,lan2019albert}, which are only available in English.
In line with the ABD, we keep our probe's hidden size the same as in the probed architectures.
Finally, we also probe an \emph{untrained} transformer model with the same architecture as BERT as a baseline.\looseness=-1

\vspace{-1pt}
\paragraph{Training.}
We train our probes with AdamW \citep{loshchilov2018decoupled} using its default hyper-parameters in PyTorch \citep{pytorch}.\footnote{The estimation of $\vent(A)$ is described in \Cref{app:unconditional_entropies}.}

\vspace{-1pt}
\paragraph{Baselines and Skylines.\footnote{We describe both approaches in more detail in \Cref{app:baselines}.}}
We contrast our attentional probe's \vinfo{} against two other values. 
First, as a baseline, we investigate a special case of our model where $\bK\!=\!\bQ$, inspired by recent work on structural probing \cite{hewitt-liang-2019-designing,hall-maudslay-etal-2020-tale,white-etal-2021-non}.
Notably, this equality leads to symmetric attention weight matrices;
by modelling the root explicitly, however, we still get a distribution over directed trees.
This baseline evaluates whether previous 
work, by over-constraining their probes, has underestimated the amount of information available to a transformer's attention mechanism.
We report this value as \textbf{structural} \binfo{}.
Second, as a skyline, we compare our attentional probe to an estimate of the true \textbf{mutual information} $\MI(\bR; \bA)$, for which we follow \citet{pimentel-etal-2020-information} in using a deep neural network (DNN) to estimate.\footnote{We note that, as demonstrated by \citet{pimentel-etal-2020-information}, the mutual information $\MI(\bR; \bA)$ is constant across contextual representations and equivalent to $\MI(\bS; \bA)$. We thus use our single best approximation of it in each language as our estimate.\looseness=-1}

\section{Results} 
We present our main results in \Cref{fig:vinfo}.
First, our probes estimate that most syntactic information is extractable in the middle layers, as previously reported by \citet{tenney_bert_2019}.
Second, \Cref{fig:vinfo} shows that a large amount of syntactic information is encoded in the representations fed to the attention heads.
Further, while we estimate close to 31 bits of information to be encoded in English, Tamil, and Basque sentences, 
we only estimate around 15 bits to be encoded in Turkish sentences; we suspect this is due to Turkish having the shortest sentences in the corpus (see \Cref{app:len} for these lengths).\looseness=-1

Third, we find that, out of the total syntactic information present in the sentences, nearly all is available to the transformer-based models under consideration. 
In English, for instance, we find 
the \vcoefficient{} of the most informative layer to be $90\%$, $82\%$, and $89\%$ in BERT, RoBERTa and ALBERT, respectively; see \Cref{tab:vcoefficients}.
This means they have access to roughly $85\%$ of all syntactic information in a sentence.
These trends are consistent across the four languages we have considered.
Notably, this is not the case for the untrained BERT representations, which suggests this structure is a byproduct of the language models'
pretraining procedures.\looseness=-1

\begin{table}
    \centering
    \begin{tabular}{lcccc}
        \toprule
        \textbf{Model} & \textbf{Layer} & $\boldsymbol{\vMI}$ & $\boldsymbol{\MI}$ & $\boldsymbol{\vcoeff}$ \\
        \midrule
        \newcolor{myblue}{BERT} & 7 & 28.0 & 31.2 & 90\% \\
        \newcolor{myred}{RoBERTa} & 9 & 25.5 & 31.2 & 82\% \\
        \newcolor{mygreen}{ALBERT} & 7 & 27.7 & 31.2  & 89\% \\
        \bottomrule
    \end{tabular}
    \vspace{-3pt}
    \caption{Maximum \vinfo{}s ($\vMI$) and \vcoefficient{}s ($\vcoeff$) estimated in English in each probed model, together with the layer in which they occur. We also display the estimated mutual information ($\MI$).\looseness=-1}
    \label{tab:vcoefficients}
    \vspace{-4pt}
\end{table}

Additionally, we find that our structural baseline 
considerably underestimates the models' potential ability to reconstruct a syntax tree; the best English structural baseline recovers only $23$ bits of information (versus $28$ bits by the attentional probe).
One can see this effect in \Cref{fig:vinfo}, where all of the structural baseline results fall beneath their corresponding attentional probe counterparts.\footnote{We provide unlabelled attachment scores in \Cref{app:uas}.}

\begin{figure}
\centering
\includegraphics[width=.9999\columnwidth]{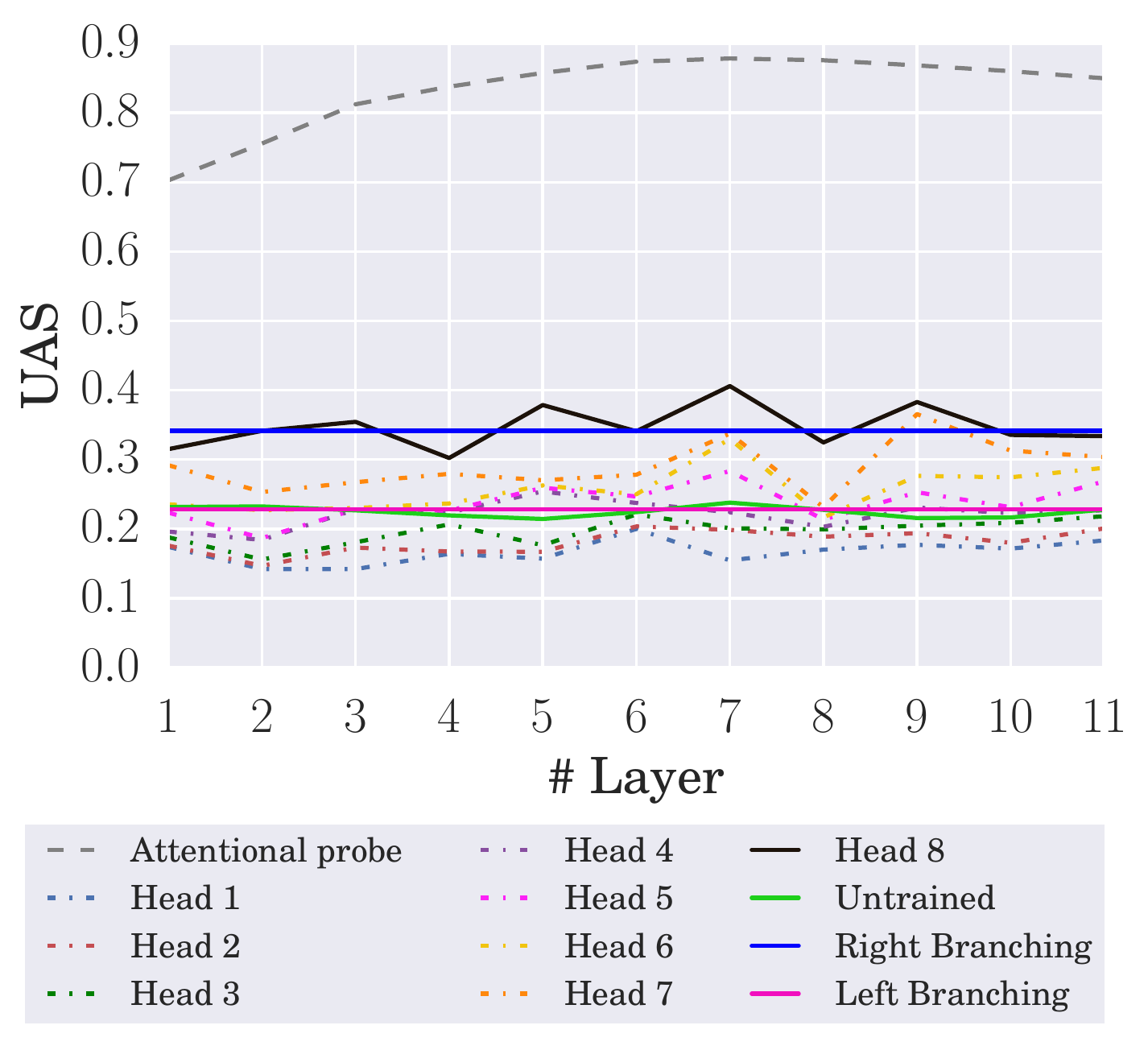}
\caption{UAS of the attentional weights computed by BERT (with its pretrained weights frozen) in English.
We also display the UAS achieved by the attentional probe, the best head per layer of an untrained BERT, and a right and left-branching baseline.\looseness=-1}
\label{fig:english_head}
\end{figure}

In a final experiment, we plug BERT's attention weights, as computed with its pretrained attention heads, directly into \Cref{eq:tree_prob} and analyse its resulting unlabelled attachment scores.
These results are presented in \Cref{fig:english_head} for English (as well as in \Cref{app:logits} for the other analysed languages).
In short, they reveal that, while attention heads \emph{could} use a large amount of syntactic information,
none of the actual heads compute weights that strongly resemble syntax trees; see \citealt{htut_etal_2019_attention} for similar results.
As BERT has 8 attention heads, however, it might be the case that the syntactic information is used in a distributed manner, with each head relying on a subset of this information (see Tab. 3 in \citealt{clark-etal-2019-bert} for results partly supporting this hypothesis).\looseness=-1

\section{Conclusions}

In this paper, we have approached probing from a new perspective.
Rather than asking how much information is encoded by the model, we ask how much information its components could extract.
We then quantify this amount using \vinfo{}.
Evaluating the attention mechanism of popular transformer language models, we find that the majority of the information about the syntax tree of a sentence is in fact extractable by the model.
This, however, is not true for randomly initialised transformer models.
Our results, thus, lead us to conclude that a transformer's training leads its attention heads to have the potential to decode syntax trees.\looseness-1

\section*{Acknowledgements}

We thank the reviewers and action editor for their helpful comments.
We also thank Kevin Du, Clara Meister, Lucas Torroba Hennigen, and Afra Amini for their feedback on this manuscript.

\section*{Limitations}

\newcommand{\suggest}[2]{{\color{red} \sout{#1}} {\color{blue}#2}}

In this paper, we propose a new principled way to choose a probe's architecture,
operationalising the question ``how much information \emph{could} a model extract from a set of representations?'' in terms of a \vinfo{}.
We note, however, that this probe design principle is only applicable to answer the specific question above.
Explicitly, we do \emph{not} answer what we believe to be the more interesting question: ``how much information \emph{does} a model actually extract from a set of representations?''
In practice, while our proposed probing method does not answer this second question, it does offer an upperbound for it; the amount of information a model could extract from a set of representations is strictly larger than the amount actually extracted.
Quantifying how tight (or loose) this upperbound is remains future work.\looseness=-1

\section*{Ethical Concerns}
We foresee no ethical concerns with this work.

\bibliography{custom}

\begin{thebibliography}{50}
\expandafter\ifx\csname natexlab\endcsname\relax\def\natexlab#1{#1}\fi

\bibitem[{Adi et~al.(2017)Adi, Kermany, Belinkov, Lavi, and
  Goldberg}]{adi2016fine}
Yossi Adi, Einat Kermany, Yonatan Belinkov, Ofer Lavi, and Yoav Goldberg. 2017.
\newblock \href {https://openreview.net/forum?id=BJh6Ztuxl} {Fine-grained
  analysis of sentence embeddings using auxiliary prediction tasks}.
\newblock In \emph{International Conference on Learning Representations}.

\bibitem[{Aduriz et~al.(2003)Aduriz, Aranzabe, Arriola, Atutxa, Diaz~de
  Ilarraza, Garmendia, and Oronoz}]{aduriz2003construction}
Itziar Aduriz, Maria~Jesus Aranzabe, Jose~Maria Arriola, Aitziber Atutxa,
  Arantza Diaz~de Ilarraza, Aitzpea Garmendia, and Maite Oronoz. 2003.
\newblock \href
  {http://ixa.si.ehu.es/sites/default/files/dokumentuak/3240/TLT2003_pdf.pdf}
  {Construction of a {B}asque dependency treebank}.
\newblock In \emph{Treebanks and Linguistic Theories}.

\bibitem[{Alain and Bengio(2016)}]{alain2016understanding}
Guillaume Alain and Yoshua Bengio. 2016.
\newblock \href {https://arxiv.org/abs/1610.01644} {Understanding intermediate
  layers using linear classifier probes}.
\newblock \emph{arXiv preprint arXiv:1610.01644}.

\bibitem[{Belinkov et~al.(2017)Belinkov, Durrani, Dalvi, Sajjad, and
  Glass}]{belinkov-etal-2017-neural}
Yonatan Belinkov, Nadir Durrani, Fahim Dalvi, Hassan Sajjad, and James Glass.
  2017.
\newblock \href {https://doi.org/10.18653/v1/P17-1080} {What do neural machine
  translation models learn about morphology?}
\newblock In \emph{Proceedings of the 55th Annual Meeting of the Association
  for Computational Linguistics (Volume 1: Long Papers)}, pages 861--872,
  Vancouver, Canada. Association for Computational Linguistics.

\bibitem[{Belinkov and Glass(2019)}]{belinkov_survey}
Yonatan Belinkov and James Glass. 2019.
\newblock \href {https://doi.org/10.1162/tacl_a_00254} {{Analysis Methods in
  Neural Language Processing: A Survey}}.
\newblock \emph{Transactions of the Association for Computational Linguistics},
  7:49--72.

\bibitem[{Bisazza and Tump(2018)}]{bisazza-tump-2018-lazy}
Arianna Bisazza and Clara Tump. 2018.
\newblock \href {https://doi.org/10.18653/v1/D18-1313} {The lazy encoder: A
  fine-grained analysis of the role of morphology in neural machine
  translation}.
\newblock In \emph{Proceedings of the 2018 Conference on Empirical Methods in
  Natural Language Processing}, pages 2871--2876, Brussels, Belgium.
  Association for Computational Linguistics.

\bibitem[{Cao et~al.(2021)Cao, Sanh, and Rush}]{cao-etal-2021-low}
Steven Cao, Victor Sanh, and Alexander Rush. 2021.
\newblock \href {https://doi.org/10.18653/v1/2021.naacl-main.74}
  {Low-complexity probing via finding subnetworks}.
\newblock In \emph{Proceedings of the 2021 Conference of the North American
  Chapter of the Association for Computational Linguistics: Human Language
  Technologies}, pages 960--966, Online. Association for Computational
  Linguistics.

\bibitem[{Clark et~al.(2019)Clark, Khandelwal, Levy, and
  Manning}]{clark-etal-2019-bert}
Kevin Clark, Urvashi Khandelwal, Omer Levy, and Christopher~D. Manning. 2019.
\newblock \href {https://doi.org/10.18653/v1/W19-4828} {What does {BERT} look
  at? {A}n analysis of {BERT}{'}s attention}.
\newblock In \emph{Proceedings of the 2019 ACL Workshop BlackboxNLP: Analyzing
  and Interpreting Neural Networks for NLP}, pages 276--286, Florence, Italy.
  Association for Computational Linguistics.

\bibitem[{Conneau et~al.(2018)Conneau, Kruszewski, Lample, Barrault, and
  Baroni}]{conneau-etal-2018-cram}
Alexis Conneau, German Kruszewski, Guillaume Lample, Lo{\"\i}c Barrault, and
  Marco Baroni. 2018.
\newblock \href {https://doi.org/10.18653/v1/P18-1198} {What you can cram into
  a single {\$}{\&}!{\#}* vector: Probing sentence embeddings for linguistic
  properties}.
\newblock In \emph{Proceedings of the 56th Annual Meeting of the Association
  for Computational Linguistics (Volume 1: Long Papers)}, pages 2126--2136,
  Melbourne, Australia. Association for Computational Linguistics.

\bibitem[{Devlin et~al.(2019)Devlin, Chang, Lee, and
  Toutanova}]{devlin-etal-2019-bert}
Jacob Devlin, Ming-Wei Chang, Kenton Lee, and Kristina Toutanova. 2019.
\newblock \href {https://doi.org/10.18653/v1/N19-1423} {{BERT}: {P}re-training
  of deep bidirectional transformers for language understanding}.
\newblock In \emph{Proceedings of the 2019 Conference of the North {A}merican
  Chapter of the Association for Computational Linguistics: Human Language
  Technologies, Volume 1 (Long and Short Papers)}, pages 4171--4186,
  Minneapolis, Minnesota. Association for Computational Linguistics.

\bibitem[{Dozat and Manning(2017)}]{dozat2016deep}
Timothy Dozat and Christopher~D. Manning. 2017.
\newblock \href {https://openreview.net/forum?id=Hk95PK9le} {Deep biaffine
  attention for neural dependency parsing}.
\newblock In \emph{International Conference on Learning Representations}.

\bibitem[{Elazar et~al.(2021)Elazar, Ravfogel, Jacovi, and
  Goldberg}]{elazar-etal-2021-amnesic}
Yanai Elazar, Shauli Ravfogel, Alon Jacovi, and Yoav Goldberg. 2021.
\newblock \href {https://doi.org/10.1162/tacl_a_00359} {Amnesic probing:
  Behavioral explanation with amnesic counterfactuals}.
\newblock \emph{Transactions of the Association for Computational Linguistics},
  9:160--175.

\bibitem[{Ettinger et~al.(2016)Ettinger, Elgohary, and
  Resnik}]{ettinger-etal-2016-probing}
Allyson Ettinger, Ahmed Elgohary, and Philip Resnik. 2016.
\newblock \href {https://doi.org/10.18653/v1/W16-2524} {Probing for semantic
  evidence of composition by means of simple classification tasks}.
\newblock In \emph{Proceedings of the 1st Workshop on Evaluating Vector-Space
  Representations for {NLP}}, pages 134--139, Berlin, Germany. Association for
  Computational Linguistics.

\bibitem[{Hewitt et~al.(2021)Hewitt, Ethayarajh, Liang, and
  Manning}]{hewitt-etal-2021-conditional}
John Hewitt, Kawin Ethayarajh, Percy Liang, and Christopher Manning. 2021.
\newblock \href {https://aclanthology.org/2021.emnlp-main.122} {Conditional
  probing: measuring usable information beyond a baseline}.
\newblock In \emph{Proceedings of the 2021 Conference on Empirical Methods in
  Natural Language Processing}, pages 1626--1639, Online and Punta Cana,
  Dominican Republic. Association for Computational Linguistics.

\bibitem[{Hewitt and Liang(2019)}]{hewitt-liang-2019-designing}
John Hewitt and Percy Liang. 2019.
\newblock \href {https://doi.org/10.18653/v1/D19-1275} {Designing and
  interpreting probes with control tasks}.
\newblock In \emph{Proceedings of the 2019 Conference on Empirical Methods in
  Natural Language Processing and the 9th International Joint Conference on
  Natural Language Processing (EMNLP-IJCNLP)}, pages 2733--2743, Hong Kong,
  China. Association for Computational Linguistics.

\bibitem[{Hewitt and Manning(2019)}]{hewitt-manning-2019-structural}
John Hewitt and Christopher~D. Manning. 2019.
\newblock \href {https://doi.org/10.18653/v1/N19-1419} {{A} structural probe
  for finding syntax in word representations}.
\newblock In \emph{Proceedings of the 2019 Conference of the North {A}merican
  Chapter of the Association for Computational Linguistics: Human Language
  Technologies, Volume 1 (Long and Short Papers)}, pages 4129--4138,
  Minneapolis, Minnesota. Association for Computational Linguistics.

\bibitem[{Htut et~al.(2019)Htut, Phang, Bordia, and
  Bowman}]{htut_etal_2019_attention}
Phu~Mon Htut, Jason Phang, Shikha Bordia, and Samuel~R. Bowman. 2019.
\newblock \href {http://arxiv.org/abs/1911.12246} {Do attention heads in {BERT}
  track syntactic dependencies?}
\newblock \emph{CoRR}, abs/1911.12246.

\bibitem[{Koo et~al.(2007)Koo, Globerson, Carreras, and
  Collins}]{koo-etal-2007-structured}
Terry Koo, Amir Globerson, Xavier Carreras, and Michael Collins. 2007.
\newblock \href {https://aclanthology.org/D07-1015} {Structured prediction
  models via the matrix-tree theorem}.
\newblock In \emph{Proceedings of the 2007 Joint Conference on Empirical
  Methods in Natural Language Processing and Computational Natural Language
  Learning ({EMNLP}-{C}o{NLL})}, pages 141--150, Prague, Czech Republic.
  Association for Computational Linguistics.

\bibitem[{Kuznetsov and Gurevych(2020)}]{kuznetsov-gurevych-2020-matter}
Ilia Kuznetsov and Iryna Gurevych. 2020.
\newblock \href {https://doi.org/10.18653/v1/2020.emnlp-main.13} {A matter of
  framing: {T}he impact of linguistic formalism on probing results}.
\newblock In \emph{Proceedings of the 2020 Conference on Empirical Methods in
  Natural Language Processing (EMNLP)}, pages 171--182, Online. Association for
  Computational Linguistics.

\bibitem[{Lan et~al.(2020)Lan, Chen, Goodman, Gimpel, Sharma, and
  Soricut}]{lan2019albert}
Zhenzhong Lan, Mingda Chen, Sebastian Goodman, Kevin Gimpel, Piyush Sharma, and
  Radu Soricut. 2020.
\newblock \href {https://openreview.net/forum?id=H1eA7AEtvS} {{ALBERT}: {A}
  lite {BERT} for self-supervised learning of language representations}.
\newblock In \emph{The 8th International Conference on Learning
  Representations}.

\bibitem[{Lasri et~al.(2022)Lasri, Pimentel, Lenci, Poibeau, and
  Cotterell}]{lasri-etal-2022-probing}
Karim Lasri, Tiago Pimentel, Alessandro Lenci, Thierry Poibeau, and Ryan
  Cotterell. 2022.
\newblock \href {https://doi.org/10.18653/v1/2022.acl-long.603} {Probing for
  the usage of grammatical number}.
\newblock In \emph{Proceedings of the 60th Annual Meeting of the Association
  for Computational Linguistics (Volume 1: Long Papers)}, pages 8818--8831,
  Dublin, Ireland. Association for Computational Linguistics.

\bibitem[{Liu et~al.(2019{\natexlab{a}})Liu, Gardner, Belinkov, Peters, and
  Smith}]{liu-etal-2019-linguistic}
Nelson~F. Liu, Matt Gardner, Yonatan Belinkov, Matthew~E. Peters, and Noah~A.
  Smith. 2019{\natexlab{a}}.
\newblock \href {https://doi.org/10.18653/v1/N19-1112} {Linguistic knowledge
  and transferability of contextual representations}.
\newblock In \emph{Proceedings of the 2019 Conference of the North {A}merican
  Chapter of the Association for Computational Linguistics: Human Language
  Technologies, Volume 1 (Long and Short Papers)}, pages 1073--1094,
  Minneapolis, Minnesota. Association for Computational Linguistics.

\bibitem[{Liu et~al.(2019{\natexlab{b}})Liu, Ott, Goyal, Du, Joshi, Chen, Levy,
  Lewis, Zettlemoyer, and Stoyanov}]{liu2019roberta}
Yinhan Liu, Myle Ott, Naman Goyal, Jingfei Du, Mandar Joshi, Danqi Chen, Omer
  Levy, Mike Lewis, Luke Zettlemoyer, and Veselin Stoyanov. 2019{\natexlab{b}}.
\newblock \href {http://arxiv.org/abs/1907.11692} {{RoBERTa}: {A} robustly
  optimized {BERT} pretraining approach}.
\newblock \emph{CoRR}, abs/1907.11692.

\bibitem[{Loshchilov and Hutter(2019)}]{loshchilov2018decoupled}
Ilya Loshchilov and Frank Hutter. 2019.
\newblock \href {https://openreview.net/forum?id=Bkg6RiCqY7} {Decoupled weight
  decay regularization}.
\newblock In \emph{International Conference on Learning Representations}.

\bibitem[{Manning et~al.(2020)Manning, Clark, Hewitt, Khandelwal, and
  Levy}]{manning2020emergent}
Christopher~D. Manning, Kevin Clark, John Hewitt, Urvashi Khandelwal, and Omer
  Levy. 2020.
\newblock \href {https://doi.org/10.1073/pnas.1907367117} {Emergent linguistic
  structure in artificial neural networks trained by self-supervision}.
\newblock \emph{Proceedings of the National Academy of Sciences},
  117(48):30046--30054.

\bibitem[{Maudslay et~al.(2020)Maudslay, Valvoda, Pimentel, Williams, and
  Cotterell}]{hall-maudslay-etal-2020-tale}
Rowan~Hall Maudslay, Josef Valvoda, Tiago Pimentel, Adina Williams, and Ryan
  Cotterell. 2020.
\newblock \href {https://doi.org/10.18653/v1/2020.acl-main.659} {A tale of a
  probe and a parser}.
\newblock In \emph{Proceedings of the 58th Annual Meeting of the Association
  for Computational Linguistics}, pages 7389--7395, Online. Association for
  Computational Linguistics.

\bibitem[{Paszke et~al.(2019)Paszke, Gross, Massa, Lerer, Bradbury, Chanan,
  Killeen, Lin, Gimelshein, Antiga, Desmaison, Kopf, Yang, DeVito, Raison,
  Tejani, Chilamkurthy, Steiner, Fang, Bai, and Chintala}]{pytorch}
Adam Paszke, Sam Gross, Francisco Massa, Adam Lerer, James Bradbury, Gregory
  Chanan, Trevor Killeen, Zeming Lin, Natalia Gimelshein, Luca Antiga, Alban
  Desmaison, Andreas Kopf, Edward Yang, Zachary DeVito, Martin Raison, Alykhan
  Tejani, Sasank Chilamkurthy, Benoit Steiner, Lu~Fang, Junjie Bai, and Soumith
  Chintala. 2019.
\newblock \href
  {http://papers.neurips.cc/paper/9015-pytorch-an-imperative-style-high-performance-deep-learning-library.pdf}
  {{PyTorch}: {A}n imperative style, high-performance deep learning library}.
\newblock In H.~Wallach, H.~Larochelle, A.~Beygelzimer, F.~d\textquotesingle
  Alch\'{e}-Buc, E.~Fox, and R.~Garnett, editors, \emph{Advances in Neural
  Information Processing Systems 32}, pages 8024--8035. Curran Associates, Inc.

\bibitem[{Pimentel and Cotterell(2021)}]{pimentel-cotterell-2021-bayesian}
Tiago Pimentel and Ryan Cotterell. 2021.
\newblock \href {https://aclanthology.org/2021.emnlp-main.229} {A {B}ayesian
  framework for information-theoretic probing}.
\newblock In \emph{Proceedings of the 2021 Conference on Empirical Methods in
  Natural Language Processing}, pages 2869--2887, Online and Punta Cana,
  Dominican Republic. Association for Computational Linguistics.

\bibitem[{Pimentel et~al.(2020{\natexlab{a}})Pimentel, Saphra, Williams, and
  Cotterell}]{pimentel-etal-2020-pareto}
Tiago Pimentel, Naomi Saphra, Adina Williams, and Ryan Cotterell.
  2020{\natexlab{a}}.
\newblock \href {https://doi.org/10.18653/v1/2020.emnlp-main.254} {{P}areto
  probing: {T}rading off accuracy for complexity}.
\newblock In \emph{Proceedings of the 2020 Conference on Empirical Methods in
  Natural Language Processing (EMNLP)}, pages 3138--3153, Online. Association
  for Computational Linguistics.

\bibitem[{Pimentel et~al.(2020{\natexlab{b}})Pimentel, Valvoda, Maudslay,
  Zmigrod, Williams, and Cotterell}]{pimentel-etal-2020-information}
Tiago Pimentel, Josef Valvoda, Rowan~Hall Maudslay, Ran Zmigrod, Adina
  Williams, and Ryan Cotterell. 2020{\natexlab{b}}.
\newblock \href {https://doi.org/10.18653/v1/2020.acl-main.420}
  {Information-theoretic probing for linguistic structure}.
\newblock In \emph{Proceedings of the 58th Annual Meeting of the Association
  for Computational Linguistics}, pages 4609--4622, Online. Association for
  Computational Linguistics.

\bibitem[{Qian et~al.(2016)Qian, Qiu, and Huang}]{qian-etal-2016-investigating}
Peng Qian, Xipeng Qiu, and Xuanjing Huang. 2016.
\newblock \href {https://doi.org/10.18653/v1/P16-1140} {Investigating language
  universal and specific properties in word embeddings}.
\newblock In \emph{Proceedings of the 54th Annual Meeting of the Association
  for Computational Linguistics (Volume 1: Long Papers)}, pages 1478--1488,
  Berlin, Germany. Association for Computational Linguistics.

\bibitem[{Ramasamy and \v{Z}abokrtsk\'{y}(2012)}]{tamil_treebank}
Loganathan Ramasamy and Zden\v{e}k \v{Z}abokrtsk\'{y}. 2012.
\newblock \href
  {http://www.lrec-conf.org/proceedings/lrec2012/summaries/456.html} {Prague
  dependency style treebank for {Tamil}}.
\newblock In \emph{Proceedings of Eighth International Conference on Language
  Resources and Evaluation ({LREC} 2012)}, pages 1888--1894, \.{I}stanbul,
  Turkey.

\bibitem[{Ravfogel et~al.(2021)Ravfogel, Prasad, Linzen, and
  Goldberg}]{ravfogel-etal-2021-counterfactual}
Shauli Ravfogel, Grusha Prasad, Tal Linzen, and Yoav Goldberg. 2021.
\newblock \href {https://doi.org/10.18653/v1/2021.conll-1.15} {Counterfactual
  interventions reveal the causal effect of relative clause representations on
  agreement prediction}.
\newblock In \emph{Proceedings of the 25th Conference on Computational Natural
  Language Learning}, pages 194--209, Online. Association for Computational
  Linguistics.

\bibitem[{Rogers et~al.(2021)Rogers, Kovaleva, and Rumshisky}]{rogers_syntax}
Anna Rogers, Olga Kovaleva, and Anna Rumshisky. 2021.
\newblock \href {https://doi.org/10.1162/tacl_a_00349} {A primer in
  {BERT}ology: {W}hat we know about how {BERT} works}.
\newblock \emph{Transactions of the Association for Computational Linguistics},
  8:842--866.

\bibitem[{Saphra and Lopez(2019)}]{saphra-lopez-2019-understanding}
Naomi Saphra and Adam Lopez. 2019.
\newblock \href {https://doi.org/10.18653/v1/N19-1329} {Understanding learning
  dynamics of language models with {SVCCA}}.
\newblock In \emph{Proceedings of the 2019 Conference of the North {A}merican
  Chapter of the Association for Computational Linguistics: Human Language
  Technologies, Volume 1 (Long and Short Papers)}, pages 3257--3267,
  Minneapolis, Minnesota. Association for Computational Linguistics.

\bibitem[{Shannon(1948)}]{shannon1948mathematical}
Claude~E. Shannon. 1948.
\newblock \href {https://doi.org/10.1002/j.1538-7305.1948.tb01338.x} {A
  mathematical theory of communication}.
\newblock \emph{The Bell System Technical Journal}, 27(3):379--423.

\bibitem[{Shi et~al.(2016)Shi, Padhi, and Knight}]{shi-etal-2016-string}
Xing Shi, Inkit Padhi, and Kevin Knight. 2016.
\newblock \href {https://doi.org/10.18653/v1/D16-1159} {Does string-based
  neural {MT} learn source syntax?}
\newblock In \emph{Proceedings of the 2016 Conference on Empirical Methods in
  Natural Language Processing}, pages 1526--1534, Austin, Texas. Association
  for Computational Linguistics.

\bibitem[{Silveira et~al.(2014)Silveira, Dozat, de~Marneffe, Bowman, Connor,
  Bauer, and Manning}]{silveira14gold}
Natalia Silveira, Timothy Dozat, Marie-Catherine de~Marneffe, Samuel Bowman,
  Miriam Connor, John Bauer, and Christopher~D. Manning. 2014.
\newblock \href {https://aclanthology.org/L14-1067/} {A gold standard
  dependency corpus for {E}nglish}.
\newblock In \emph{Proceedings of the Ninth International Conference on
  Language Resources and Evaluation (LREC-2014)}.

\bibitem[{Sulubacak et~al.(2016)Sulubacak, Gokirmak, Tyers, {\c{C}}{\"o}ltekin,
  Nivre, and Eryi{\u{g}}it}]{sulubacak-etal-2016-universal}
Umut Sulubacak, Memduh Gokirmak, Francis Tyers, {\c{C}}a{\u{g}}r{\i}
  {\c{C}}{\"o}ltekin, Joakim Nivre, and G{\"u}l{\c{s}}en Eryi{\u{g}}it. 2016.
\newblock \href {https://aclanthology.org/C16-1325} {{U}niversal {D}ependencies
  for {T}urkish}.
\newblock In \emph{Proceedings of {COLING} 2016, the 26th International
  Conference on Computational Linguistics: Technical Papers}, pages 3444--3454,
  Osaka, Japan. The COLING 2016 Organizing Committee.

\bibitem[{Tenney et~al.(2019)Tenney, Das, and Pavlick}]{tenney_bert_2019}
Ian Tenney, Dipanjan Das, and Ellie Pavlick. 2019.
\newblock \href {https://doi.org/10.18653/v1/P19-1452} {{BERT} rediscovers the
  classical {NLP} pipeline}.
\newblock In \emph{Proceedings of the 57th Annual Meeting of the Association
  for Computational Linguistics}, pages 4593--4601, Florence, Italy.
  Association for Computational Linguistics.

\bibitem[{Tutte(1984)}]{tutte1984graph}
W.~T. Tutte. 1984.
\newblock \href
  {https://onlinelibrary.wiley.com/doi/abs/10.1002/net.3230160110} {\emph{Graph
  Theory}}.
\newblock Addison-Wesley Publishing Company.

\bibitem[{Vaswani et~al.(2017)Vaswani, Shazeer, Parmar, Uszkoreit, Jones,
  Gomez, Kaiser, and Polosukhin}]{vaswani2017attention}
Ashish Vaswani, Noam Shazeer, Niki Parmar, Jakob Uszkoreit, Llion Jones,
  Aidan~N. Gomez, \L{}ukasz Kaiser, and Illia Polosukhin. 2017.
\newblock \href
  {https://proceedings.neurips.cc/paper/2017/file/3f5ee243547dee91fbd053c1c4a845aa-Paper.pdf}
  {Attention is all you need}.
\newblock In \emph{Advances in Neural Information Processing Systems},
  volume~30. Curran Associates, Inc.

\bibitem[{Vig and Belinkov(2019)}]{vig-belinkov-2019-analyzing}
Jesse Vig and Yonatan Belinkov. 2019.
\newblock \href {https://doi.org/10.18653/v1/W19-4808} {Analyzing the structure
  of attention in a transformer language model}.
\newblock In \emph{Proceedings of the 2019 ACL Workshop BlackboxNLP: Analyzing
  and Interpreting Neural Networks for NLP}, pages 63--76, Florence, Italy.
  Association for Computational Linguistics.

\bibitem[{Voita and Titov(2020)}]{voita-titov-2020-information}
Elena Voita and Ivan Titov. 2020.
\newblock \href {https://doi.org/10.18653/v1/2020.emnlp-main.14}
  {Information-theoretic probing with minimum description length}.
\newblock In \emph{Proceedings of the 2020 Conference on Empirical Methods in
  Natural Language Processing (EMNLP)}, pages 183--196, Online. Association for
  Computational Linguistics.

\bibitem[{White et~al.(2021)White, Pimentel, Saphra, and
  Cotterell}]{white-etal-2021-non}
Jennifer~C. White, Tiago Pimentel, Naomi Saphra, and Ryan Cotterell. 2021.
\newblock \href {https://doi.org/10.18653/v1/2021.naacl-main.12} {A non-linear
  structural probe}.
\newblock In \emph{Proceedings of the 2021 Conference of the North American
  Chapter of the Association for Computational Linguistics: Human Language
  Technologies}, pages 132--138, Online. Association for Computational
  Linguistics.

\bibitem[{Wolf et~al.(2020)Wolf, Debut, Sanh, Chaumond, Delangue, Moi, Cistac,
  Rault, Louf, Funtowicz, Davison, Shleifer, von Platen, Ma, Jernite, Plu, Xu,
  Le~Scao, Gugger, Drame, Lhoest, and Rush}]{wolf-etal-2020-transformers}
Thomas Wolf, Lysandre Debut, Victor Sanh, Julien Chaumond, Clement Delangue,
  Anthony Moi, Pierric Cistac, Tim Rault, Remi Louf, Morgan Funtowicz, Joe
  Davison, Sam Shleifer, Patrick von Platen, Clara Ma, Yacine Jernite, Julien
  Plu, Canwen Xu, Teven Le~Scao, Sylvain Gugger, Mariama Drame, Quentin Lhoest,
  and Alexander Rush. 2020.
\newblock \href {https://doi.org/10.18653/v1/2020.emnlp-demos.6} {Transformers:
  State-of-the-art natural language processing}.
\newblock In \emph{Proceedings of the 2020 Conference on Empirical Methods in
  Natural Language Processing: System Demonstrations}, pages 38--45, Online.
  Association for Computational Linguistics.

\bibitem[{Xu et~al.(2020)Xu, Zhao, Song, Stewart, and Ermon}]{xu2020theory}
Yilun Xu, Shengjia Zhao, Jiaming Song, Russell Stewart, and Stefano Ermon.
  2020.
\newblock \href {https://openreview.net/forum?id=r1eBeyHFDH} {A theory of
  usable information under computational constraints}.
\newblock In \emph{International Conference on Learning Representations}.

\bibitem[{Zeman et~al.(2020)Zeman, Nivre, Abrams, Ackermann, Aepli, Agi{\'c},
  Ahrenberg, Ajede, Aleksandravi{\v c}i{\=u}t{\.e}, Antonsen, Aplonova, Aquino,
  Aranzabe, Arutie, Asahara, Ateyah, Atmaca, Attia, Atutxa, Augustinus,
  Badmaeva, Ballesteros, Banerjee, Bank, Barbu~Mititelu, Basmov, Batchelor,
  Bauer, Bengoetxea, Berzak, Bhat, Bhat, Biagetti, Bick, Bielinskien{\.e},
  Blokland, Bobicev, Boizou, Borges~V{\"o}lker, B{\"o}rstell, Bosco, Bouma,
  Bowman, Boyd, Brokait{\.e}, Burchardt, Candito, Caron, Caron, Cavalcanti,
  Cebiro{\u g}lu~Eryi{\u g}it, Cecchini, Celano, {\v C}{\'e}pl{\"o}, Cetin,
  Chalub, Chi, Choi, Cho, Chun, Cignarella, Cinkov{\'a}, Collomb, {\c
  C}{\"o}ltekin, Connor, Courtin, Davidson, de~Marneffe, de~Paiva, de~Souza,
  Diaz~de Ilarraza, Dickerson, Dione, Dirix, Dobrovoljc, Dozat, Droganova,
  Dwivedi, Eckhoff, Eli, Elkahky, Ephrem, Erina, Erjavec, Etienne, Evelyn,
  Farkas, Fernandez~Alcalde, Foster, Freitas, Fujita, Gajdo{\v s}ov{\'a},
  Galbraith, Garcia, G{\"a}rdenfors, Garza, Gerdes, Ginter, Goenaga, Gojenola,
  G{\"o}k{\i}rmak, Goldberg, G{\'o}mez~Guinovart, Gonz{\'a}lez~Saavedra,
  Grici{\=u}t{\.e}, Grioni, Grobol, Gr{\= u}z{\={\i}}tis, Guillaume,
  Guillot-Barbance, G{\"u}ng{\"o}r, Habash, Haji{\v c}, Haji{\v c}~jr.,
  H{\"a}m{\"a}l{\"a}inen, H{\`a}~M{\~y}, Han, Harris, Haug, Heinecke, Hellwig,
  Hennig, Hladk{\'a}, Hlav{\'a}{\v c}ov{\'a}, Hociung, Hohle, Hwang, Ikeda,
  Ion, Irimia, Ishola, Jel{\'{\i}}nek, Johannsen, J{\'o}nsd{\'o}ttir,
  J{\o}rgensen, Juutinen, Ka{\c s}{\i}kara, Kaasen, Kabaeva, Kahane, Kanayama,
  Kanerva, Katz, Kayadelen, Kenney, Kettnerov{\'a}, Kirchner, Klementieva,
  K{\"o}hn, K{\"o}ksal, Kopacewicz, Korkiakangas, Kotsyba, Kovalevskait{\.e},
  Krek, Kwak, Laippala, Lambertino, Lam, Lando, Larasati, Lavrentiev, Lee,
  L{\^e}~H{\`{\^o}}ng, Lenci, Lertpradit, Leung, Levina, Li, Li, Li, Lim, Li,
  Ljube{\v s}i{\'c}, Loginova, Lyashevskaya, Lynn, Macketanz, Makazhanov,
  Mandl, Manning, Manurung, M{\u a}r{\u a}nduc, Mare{\v c}ek, Marheinecke,
  Mart{\'{\i}}nez~Alonso, Martins, Ma{\v s}ek, Matsuda, Matsumoto, {McDonald},
  {McGuinness}, Mendon{\c c}a, Miekka, Misirpashayeva, Missil{\"a}, Mititelu,
  Mitrofan, Miyao, Montemagni, More, Moreno~Romero, Mori, Morioka, Mori, Moro,
  Mortensen, Moskalevskyi, Muischnek, Munro, Murawaki, M{\"u}{\"u}risep,
  Nainwani, Navarro~Hor{\~n}iacek, Nedoluzhko, Ne{\v s}pore-B{\=e}rzkalne,
  Nguy{\~{\^e}}n~Th{\d i}, Nguy{\~{\^e}}n Th{\d i}~Minh, Nikaido, Nikolaev,
  Nitisaroj, Nurmi, Ojala, Ojha, Ol{\'u}{\`o}kun, Omura, Onwuegbuzia, Osenova,
  {\"O}stling, {\O}vrelid, {\"O}zate{\c s}, {\"O}zg{\"u}r, {\"O}zt{\"u}rk~Ba{\c
  s}aran, Partanen, Pascual, Passarotti, Patejuk, Paulino-Passos,
  Peljak-{\L}api{\'n}ska, Peng, Perez, Perrier, Petrova, Petrov, Phelan,
  Piitulainen, Pirinen, Pitler, Plank, Poibeau, Ponomareva, Popel, Pretkalni{\c
  n}a, Pr{\'e}vost, Prokopidis, Przepi{\'o}rkowski, Puolakainen, Pyysalo, Qi,
  R{\"a}{\"a}bis, Rademaker, Ramasamy, Rama, Ramisch, Ravishankar, Real,
  Rebeja, Reddy, Rehm, Riabov, Rie{\ss}ler, Rimkut{\.e}, Rinaldi, Rituma,
  Rocha, Romanenko, Rosa, Roșca, Rovati, Rudina, Rueter, Sadde, Sagot, Saleh,
  Salomoni, Samard{\v z}i{\'c}, Samson, Sanguinetti, S{\"a}rg, Saul{\={\i}}te,
  Sawanakunanon, Scarlata, Schneider, Schuster, Seddah, Seeker, Seraji, Shen,
  Shimada, Shirasu, Shohibussirri, Sichinava, Silveira, Silveira, Simi,
  Simionescu, Simk{\'o}, {\v S}imkov{\'a}, Simov, Skachedubova, Smith,
  Soares-Bastos, Spadine, Stella, Straka, Strnadov{\'a}, Suhr, Sulubacak,
  Suzuki, Sz{\'a}nt{\'o}, Taji, Takahashi, Tamburini, Tanaka, Tella, Tellier,
  Thomas, Torga, Toska, Trosterud, Trukhina, Tsarfaty, T{\"u}rk, Tyers,
  Uematsu, Untilov, Ure{\v s}ov{\'a}, Uria, Uszkoreit, Utka, Vajjala, van
  Niekerk, van Noord, Varga, Villemonte de~la Clergerie, Vincze, Wakasa,
  Wallin, Walsh, Wang, Washington, Wendt, Widmer, Williams, Wir{\'e}n, Wittern,
  Woldemariam, Wong, Wr{\'o}blewska, Yako, Yamashita, Yamazaki, Yan, Yasuoka,
  Yavrumyan, Yu, {\v Z}abokrtsk{\'y}, Zeldes, Zhu, and Zhuravleva}]{ud-2.6}
Daniel Zeman, Joakim Nivre, Mitchell Abrams, Elia Ackermann, No{\"e}mi Aepli,
  {\v Z}eljko Agi{\'c}, Lars Ahrenberg, Chika~Kennedy Ajede, Gabriel{\.e}
  Aleksandravi{\v c}i{\=u}t{\.e}, Lene Antonsen, Katya Aplonova, Angelina
  Aquino, Maria~Jesus Aranzabe, Gashaw Arutie, Masayuki Asahara, Luma Ateyah,
  Furkan Atmaca, Mohammed Attia, Aitziber Atutxa, Liesbeth Augustinus, Elena
  Badmaeva, Miguel Ballesteros, Esha Banerjee, Sebastian Bank, Verginica
  Barbu~Mititelu, Victoria Basmov, Colin Batchelor, John Bauer, Kepa
  Bengoetxea, Yevgeni Berzak, Irshad~Ahmad Bhat, Riyaz~Ahmad Bhat, Erica
  Biagetti, Eckhard Bick, Agn{\.e} Bielinskien{\.e}, Rogier Blokland, Victoria
  Bobicev, Lo{\"{\i}}c Boizou, Emanuel Borges~V{\"o}lker, Carl B{\"o}rstell,
  Cristina Bosco, Gosse Bouma, Sam Bowman, Adriane Boyd, Kristina Brokait{\.e},
  Aljoscha Burchardt, Marie Candito, Bernard Caron, Gauthier Caron, Tatiana
  Cavalcanti, G{\"u}l{\c s}en Cebiro{\u g}lu~Eryi{\u g}it, Flavio~Massimiliano
  Cecchini, Giuseppe G.~A. Celano, Slavom{\'{\i}}r {\v C}{\'e}pl{\"o}, Savas
  Cetin, Fabricio Chalub, Ethan Chi, Jinho Choi, Yongseok Cho, Jayeol Chun,
  Alessandra~T. Cignarella, Silvie Cinkov{\'a}, Aur{\'e}lie Collomb, {\c C}a{\u
  g}r{\i} {\c C}{\"o}ltekin, Miriam Connor, Marine Courtin, Elizabeth Davidson,
  Marie-Catherine de~Marneffe, Valeria de~Paiva, Elvis de~Souza, Arantza
  Diaz~de Ilarraza, Carly Dickerson, Bamba Dione, Peter Dirix, Kaja Dobrovoljc,
  Timothy Dozat, Kira Droganova, Puneet Dwivedi, Hanne Eckhoff, Marhaba Eli,
  Ali Elkahky, Binyam Ephrem, Olga Erina, Toma{\v z} Erjavec, Aline Etienne,
  Wograine Evelyn, Rich{\'a}rd Farkas, Hector Fernandez~Alcalde, Jennifer
  Foster, Cl{\'a}udia Freitas, Kazunori Fujita, Katar{\'{\i}}na Gajdo{\v
  s}ov{\'a}, Daniel Galbraith, Marcos Garcia, Moa G{\"a}rdenfors, Sebastian
  Garza, Kim Gerdes, Filip Ginter, Iakes Goenaga, Koldo Gojenola, Memduh
  G{\"o}k{\i}rmak, Yoav Goldberg, Xavier G{\'o}mez~Guinovart, Berta
  Gonz{\'a}lez~Saavedra, Bernadeta Grici{\=u}t{\.e}, Matias Grioni, Lo{\"{\i}}c
  Grobol, Normunds Gr{\= u}z{\={\i}}tis, Bruno Guillaume, C{\'e}line
  Guillot-Barbance, Tunga G{\"u}ng{\"o}r, Nizar Habash, Jan Haji{\v c}, Jan
  Haji{\v c}~jr., Mika H{\"a}m{\"a}l{\"a}inen, Linh H{\`a}~M{\~y}, Na-Rae Han,
  Kim Harris, Dag Haug, Johannes Heinecke, Oliver Hellwig, Felix Hennig,
  Barbora Hladk{\'a}, Jaroslava Hlav{\'a}{\v c}ov{\'a}, Florinel Hociung,
  Petter Hohle, Jena Hwang, Takumi Ikeda, Radu Ion, Elena Irimia, {\d
  O}l{\'a}j{\'{\i}}d{\'e} Ishola, Tom{\'a}{\v s} Jel{\'{\i}}nek, Anders
  Johannsen, Hildur J{\'o}nsd{\'o}ttir, Fredrik J{\o}rgensen, Markus Juutinen,
  H{\"u}ner Ka{\c s}{\i}kara, Andre Kaasen, Nadezhda Kabaeva, Sylvain Kahane,
  Hiroshi Kanayama, Jenna Kanerva, Boris Katz, Tolga Kayadelen, Jessica Kenney,
  V{\'a}clava Kettnerov{\'a}, Jesse Kirchner, Elena Klementieva, Arne K{\"o}hn,
  Abdullatif K{\"o}ksal, Kamil Kopacewicz, Timo Korkiakangas, Natalia Kotsyba,
  Jolanta Kovalevskait{\.e}, Simon Krek, Sookyoung Kwak, Veronika Laippala,
  Lorenzo Lambertino, Lucia Lam, Tatiana Lando, Septina~Dian Larasati, Alexei
  Lavrentiev, John Lee, Phương L{\^e}~H{\`{\^o}}ng, Alessandro Lenci, Saran
  Lertpradit, Herman Leung, Maria Levina, Cheuk~Ying Li, Josie Li, Keying Li,
  {KyungTae} Lim, Yuan Li, Nikola Ljube{\v s}i{\'c}, Olga Loginova, Olga
  Lyashevskaya, Teresa Lynn, Vivien Macketanz, Aibek Makazhanov, Michael Mandl,
  Christopher Manning, Ruli Manurung, C{\u a}t{\u a}lina M{\u a}r{\u a}nduc,
  David Mare{\v c}ek, Katrin Marheinecke, H{\'e}ctor Mart{\'{\i}}nez~Alonso,
  Andr{\'e} Martins, Jan Ma{\v s}ek, Hiroshi Matsuda, Yuji Matsumoto, Ryan
  {McDonald}, Sarah {McGuinness}, Gustavo Mendon{\c c}a, Niko Miekka, Margarita
  Misirpashayeva, Anna Missil{\"a}, C{\u a}t{\u a}lin Mititelu, Maria Mitrofan,
  Yusuke Miyao, Simonetta Montemagni, Amir More, Laura Moreno~Romero,
  Keiko~Sophie Mori, Tomohiko Morioka, Shinsuke Mori, Shigeki Moro, Bjartur
  Mortensen, Bohdan Moskalevskyi, Kadri Muischnek, Robert Munro, Yugo Murawaki,
  Kaili M{\"u}{\"u}risep, Pinkey Nainwani, Juan~Ignacio Navarro~Hor{\~n}iacek,
  Anna Nedoluzhko, Gunta Ne{\v s}pore-B{\=e}rzkalne, Lương
  Nguy{\~{\^e}}n~Th{\d i}, Huy{\`{\^e}}n Nguy{\~{\^e}}n Th{\d i}~Minh,
  Yoshihiro Nikaido, Vitaly Nikolaev, Rattima Nitisaroj, Hanna Nurmi, Stina
  Ojala, Atul~Kr. Ojha, Ad{\'e}day{\d o} Ol{\'u}{\`o}kun, Mai Omura, Emeka
  Onwuegbuzia, Petya Osenova, Robert {\"O}stling, Lilja {\O}vrelid, {\c
  S}aziye~Bet{\"u}l {\"O}zate{\c s}, Arzucan {\"O}zg{\"u}r, Balk{\i}z
  {\"O}zt{\"u}rk~Ba{\c s}aran, Niko Partanen, Elena Pascual, Marco Passarotti,
  Agnieszka Patejuk, Guilherme Paulino-Passos, Angelika Peljak-{\L}api{\'n}ska,
  Siyao Peng, Cenel-Augusto Perez, Guy Perrier, Daria Petrova, Slav Petrov,
  Jason Phelan, Jussi Piitulainen, Tommi~A Pirinen, Emily Pitler, Barbara
  Plank, Thierry Poibeau, Larisa Ponomareva, Martin Popel, Lauma Pretkalni{\c
  n}a, Sophie Pr{\'e}vost, Prokopis Prokopidis, Adam Przepi{\'o}rkowski, Tiina
  Puolakainen, Sampo Pyysalo, Peng Qi, Andriela R{\"a}{\"a}bis, Alexandre
  Rademaker, Loganathan Ramasamy, Taraka Rama, Carlos Ramisch, Vinit
  Ravishankar, Livy Real, Petru Rebeja, Siva Reddy, Georg Rehm, Ivan Riabov,
  Michael Rie{\ss}ler, Erika Rimkut{\.e}, Larissa Rinaldi, Laura Rituma, Luisa
  Rocha, Mykhailo Romanenko, Rudolf Rosa, Valentin Roșca, Davide Rovati, Olga
  Rudina, Jack Rueter, Shoval Sadde, Beno{\^{\i}}t Sagot, Shadi Saleh, Alessio
  Salomoni, Tanja Samard{\v z}i{\'c}, Stephanie Samson, Manuela Sanguinetti,
  Dage S{\"a}rg, Baiba Saul{\={\i}}te, Yanin Sawanakunanon, Salvatore Scarlata,
  Nathan Schneider, Sebastian Schuster, Djam{\'e} Seddah, Wolfgang Seeker,
  Mojgan Seraji, Mo~Shen, Atsuko Shimada, Hiroyuki Shirasu, Muh Shohibussirri,
  Dmitry Sichinava, Aline Silveira, Natalia Silveira, Maria Simi, Radu
  Simionescu, Katalin Simk{\'o}, M{\'a}ria {\v S}imkov{\'a}, Kiril Simov, Maria
  Skachedubova, Aaron Smith, Isabela Soares-Bastos, Carolyn Spadine, Antonio
  Stella, Milan Straka, Jana Strnadov{\'a}, Alane Suhr, Umut Sulubacak, Shingo
  Suzuki, Zsolt Sz{\'a}nt{\'o}, Dima Taji, Yuta Takahashi, Fabio Tamburini,
  Takaaki Tanaka, Samson Tella, Isabelle Tellier, Guillaume Thomas, Liisi
  Torga, Marsida Toska, Trond Trosterud, Anna Trukhina, Reut Tsarfaty, Utku
  T{\"u}rk, Francis Tyers, Sumire Uematsu, Roman Untilov, Zde{\v n}ka Ure{\v
  s}ov{\'a}, Larraitz Uria, Hans Uszkoreit, Andrius Utka, Sowmya Vajjala,
  Daniel van Niekerk, Gertjan van Noord, Viktor Varga, Eric Villemonte de~la
  Clergerie, Veronika Vincze, Aya Wakasa, Lars Wallin, Abigail Walsh, Jing~Xian
  Wang, Jonathan~North Washington, Maximilan Wendt, Paul Widmer, Seyi Williams,
  Mats Wir{\'e}n, Christian Wittern, Tsegay Woldemariam, Tak-sum Wong, Alina
  Wr{\'o}blewska, Mary Yako, Kayo Yamashita, Naoki Yamazaki, Chunxiao Yan,
  Koichi Yasuoka, Marat~M. Yavrumyan, Zhuoran Yu, Zden{\v e}k {\v
  Z}abokrtsk{\'y}, Amir Zeldes, Hanzhi Zhu, and Anna Zhuravleva. 2020.
\newblock \href {http://hdl.handle.net/11234/1-3226} {Universal dependencies
  2.6}.
\newblock {LINDAT}/{CLARIAH}-{CZ} digital library at the Institute of Formal
  and Applied Linguistics ({{\'U}FAL}), Faculty of Mathematics and Physics,
  Charles University.

\bibitem[{Zmigrod et~al.(2020)Zmigrod, Vieira, and
  Cotterell}]{zmigrod-etal-2020-please}
Ran Zmigrod, Tim Vieira, and Ryan Cotterell. 2020.
\newblock \href {https://doi.org/10.18653/v1/2020.emnlp-main.390} {Please mind
  the root: {D}ecoding arborescences for dependency parsing}.
\newblock In \emph{Proceedings of the 2020 Conference on Empirical Methods in
  Natural Language Processing (EMNLP)}, pages 4809--4819, Online. Association
  for Computational Linguistics.

\bibitem[{Zmigrod et~al.(2021)Zmigrod, Vieira, and
  Cotterell}]{zmigrod-etal-2021-efficient-computation}
Ran Zmigrod, Tim Vieira, and Ryan Cotterell. 2021.
\newblock \href {https://doi.org/10.1162/tacl_a_00391} {Efficient computation
  of expectations under spanning tree distributions}.
\newblock \emph{Transactions of the Association for Computational Linguistics},
  9:675--690.

\end{thebibliography}
\bibliographystyle{acl_natbib}

\clearpage
\appendix

\section{More on the \binfo{}}

\subsection{Probing as approximating \binfo{}} \label{sec:vinfo_and_probing}

In this section, we make a similar argument to \citeposs{hewitt-etal-2021-conditional}, who first pointed out the equivalence between the goals of probing and estimating a \vinfo{}.
When probing for some information, we typically train a probabilistic classifier $\qtheta(\ba \!\mid\! \br)$ (with parameters $\btheta$) to approximate a target probability distribution $p(\ba \!\mid\! \br)$.
We do this by using an empirical cross-entropy loss function
\begin{equation}\label{eq:loss}
    \loss(\trainset; \btheta) \defeq \sum_{(\br, \ba) \in \trainset{}} \log \frac{1}{\qtheta(\ba \!\mid\! \br)}
\end{equation}
where $\trainset$ is a training set composed of $(\br, \ba)$ pairs, which are assumed to be sampled from the true distribution $p(\br, \ba)$.
Further, we usually have access to a development set $\devset$, on which we estimate this same loss $\loss(\devset; \btheta)$ and which we use to avoid overfitting.
Together, these steps aim at making $\qtheta(\ba \!\mid\! \br)$ approximate the distribution which minimises the true cross-entropy
\begin{equation} \label{eq:condition_xent}
\xent(\bA \mid \bR) \defeq \int\limits_{\calR}\, \sum_{\ba \in \calA} p(\br, \ba) \log \frac{1}{\qtheta(\ba \!\mid\! \br)} \dr
\end{equation}
Minimising this cross-entropy is equivalent to finding the $q(\ba \!\mid\! \br) \in \calV$ which minimises the conditional $\calV$-entropy in \Cref{eq:condition_vent}.
Furthermore, since $\vent(\bA)$ is constant with respect to the representations $\bR$, this is also equivalent (up to an additive constant) to estimating the \vinfo{} in \Cref{eq:vinfo}---where $\calV$ is defined by our choice of architecture for the probing classifier.

\subsection{On $\bm{\calV}$, Expressivity and Learnability} \label{sec:vinfo_vs_learnability}

Ideally, a trained probe $\qtheta(\br \mid \ba)$ would converge to the infimum $q \in \calV$ from \Cref{eq:condition_vent}. In practice, however, limitations on the dataset size and optimisation algorithms may lead to poor approximations.
Moreover, even with a trained probe, we still cannot compute \Cref{eq:condition_xent}, but must instead empirically approximate it with a test set and the loss function in \Cref{eq:loss}.
We can thus decompose our actually measured value into four terms
\begin{align}
    \loss(\testset; \btheta) = \ent(\bA \mid \bR) + \epsilon_1 + \epsilon_2 + \epsilon_3
\end{align}
where
\begin{align}
     \epsilon_1 &\defeq \underbrace{\vent(\bA \mid \bR) - \ent(\bA \mid \bR)}_{\mathtt{Expressivity\ Constraint}} \\
     \epsilon_2 &\defeq \underbrace{\xent(\bA \mid \bR) - \vent(\bA \mid \bR)}_{\mathtt{Training\ Constraint}} \\
     \epsilon_3 &\defeq \underbrace{\loss(\testset; \btheta) - \xent(\bA \mid \bR)}_{\mathtt{Measurement\ Error}}
\end{align}
Given a large enough testset, $\epsilon_3$ should be roughly zero, as the empirical loss in \Cref{eq:loss} is an unbiased estimator of the cross-entropy in \Cref{eq:condition_xent}.
This leaves $\epsilon_1$ and $\epsilon_2$.
While $\epsilon_1$ is intentionally imposed by the choice of $\calV$, which defines the expressivity constraints on the structure of the learned information extractors, $\epsilon_2$ is a byproduct of multiple factors: $\calV$ itself, the optimisation algorithm and both the train and devset sizes.

Analysing the \vinfo{} of a very expressive variational family may thus be vacuous, as we may expect $\epsilon_1$ to be relatively small compared to $\epsilon_2$; this would likely be the case for a $\calV$ resembling the entire BERT architecture.\footnote{In these scenarios, an information-theoretic measure that accounts for training set sizes might be more meaningful, such as the Bayesian information \citep{pimentel-cotterell-2021-bayesian}.}
For smaller variational families, however, such as the ones we explore here, we can expect our learning procedures to be well behaved and for $\epsilon_2$ to be relatively small.

\section{Inverse Ablation Perspective}

One could view our work as a reversed ablation study.
In a typical ablation experiment a component of a model is removed to observe its effect on the functioning of the entire model.
The idea is that the observed difference in performance of the model will indicate the relative importance of the component.
However, with ablation it is impossible to tell what role the component plays in solving the target task.
In comparison, we freeze the entire model up to a particular component we are interested in.
Instead of asking how important the component is to the overall goal of the model, we ask how good it is at a task we believe is important towards achieving such goal.

\section{Baseline and Skyline} \label{app:baselines}

\subsection{Structural Baseline}

\citet{hewitt-manning-2019-structural} propose the structural probe to investigate the encoding of syntactic structure in contextual representations.
Intuitively, they probe to which extent they can reconstruct a sentence's syntactic tree purely from the distance between contextual representations $\br$. 
Instead of learning separate query $\bQ$ and key $\bK$ matrices as we do, however, they limit themselves to a single projection matrix $\bB \in \R^{d_2 \times d_1}$.
Their probe can thus be written as
\begin{equation} \label{eq:structural_1}
    \alpha_{ij} = (\bB \br_i)^\intercal\, \bB \br_j
\end{equation}
Since we want to train the structural probe with the same cross-entropy parsing loss as our attentional probe, we softmax its distances
\begin{equation} \label{eq:structural_softmax}
\nbw_{ij} = \frac{e^{\alpha_{ij}}}{\sum\limits_{1 \le j' \le |\bs|} e^{\alpha_{ij'}}}
\end{equation}
making it similar to \citeposs{white-etal-2021-non} non-linear structural probe.
This is necessary because the MTT we use to compute the denominator in \Cref{eq:tree_prob} assumes non-negative inputs.
We then train it with the same loss function as our proposed attentional probe, also making it similar to \citeposs{hall-maudslay-etal-2020-tale} structural parser.
In practice, thus, our structural baseline's implementation can be seen as a non-linear structural parser.

\subsection{DNN Parser}

To approximate the true mutual information $\MI(\bR; \bA)$, we follow \citet{pimentel-etal-2020-information} in using more powerful feed forward neural network probes.
Specifically, we rely on a variant of \citeposs{dozat2016deep} parser.
We first use two multi-layer perceptrons (MLP), one for the dependent and one for the head token in a dependency arc
\begin{equation} \label{eq:mlp_1}
    \br_i' = \MLP(\br_i), \qquad
    \br_j' = \MLP(\br_j)
\end{equation}
These MLP's are composed of a number of linear transformations, interweaved with ReLU non-linearities and dropout layers.
We then feed both these transformed representations into a biaffine transformation to get the dependency logits
\begin{equation} \label{eq:mlp_2}
    \alpha_{ij} = \br_i^{\prime\,\intercal}\, \bW\, \br'_j
\end{equation}
Finally, we again make these values non-negative by softmaxing them
\begin{equation} \label{eq:mlp_softmax}
\nbw_{ij} = \frac{e^{\alpha_{ij}}}{\sum\limits_{1 \le j' \le |\bs|} e^{\alpha_{ij'}}}
\end{equation}
We train this model with the same cross-entropy loss function as our proposed attentional probe.
To choose the hyper-parameters of this model's MLP we use random search, training 50 independent models.
We random search for the number of layers in $\{0, 1, 2\}$, dropout in $[0.0, 0.5]$, and the hidden size in $[32; 512]$.
Furthermore, we note that, as demonstrated by \citet{pimentel-etal-2020-information}, the mutual information $\MI(\bR; \bA)$ is constant across contextual representations and equivalent to $\MI(\bS; \bA)$, where $\bS$ is a random variable representing the original input sentence.
We thus use our single best approximation of it in each language as our estimate.\looseness=-1

\section{Unconditional Entropy Parser} \label{app:unconditional_entropies}

We still need to estimate the unconditional entropies $\vent(\bA)$.
As these unconditional entropies are not conditioned on anything, however, we cannot estimate them using the previous parsers directly.
Specifically, the representations $\br_i$ and $\br_j$ in \Cref{eq:attention_head,eq:structural_1,eq:mlp_1} cannot be used.
We sidestep this issue by dropping our contextual representations from these equations and using position embeddings in their place.
Importantly, these position embeddings do not depend on the input sentences.
In short, we compute these equations as
\begin{align} \label{eq:position_embs}
    &\alpha_{ij} = (\bK \bp_i)^\intercal\, \bQ \bp_j 
    \,\,\,\,\,\,\,\,\,\,
    \textcolor{mygray}{(\mathtt{attentional})}\\
    &\alpha_{ij} = (\bB \bp_i)^\intercal\, \bB \bp_j 
    \,\,\qquad
    \textcolor{mygray}{(\mathtt{structural})}\\
    &\br_i' = \MLP(\bp_i),\,\, \br_j' = \MLP(\bp_j)  
    \,\,\,
    \textcolor{mygray}{(\mathtt{DNN})}
\end{align}
where $\bp_i \in \R^{d_1}$ is a randomly initialised position embedding and is trained with the rest of the probe.\looseness=-1

\section{Extra Information about Training}

We use the \texttt{base} version of all our analysed pretrained models \cite[taken from the transformers library][]{wolf-etal-2020-transformers}.
We train the model with a batch size of 2048, evaluate the model every 100 batches, and stop training when the model does not improve over 10 consecutive evaluations.
Both the attentional and structural probes are trained with a dropout of $0.2$ (applied both on the raw input representations and on the key and query representations before being multiplied together) and with a hidden size (i.e. $d_2$) of 64---this is the size of the query, and key representations in both BERT, RoBERTa and ALBERT.
As our data, we used the treebanks: English EWT \citep{silveira14gold}; Basque BDT \citep{aduriz2003construction}; Turkish IMST \citep{sulubacak-etal-2016-universal}; Tamil TTB \citep{tamil_treebank}.

\clearpage
\onecolumn

\section{UAS Results}\label{app:uas}

\newcommand{\appendixfigsizes}{.85\textwidth}

\begin{figure}[h]
    \centering
     \begin{subfigure}[b]{0.5\textwidth}
         \centering
         \includegraphics[width=\appendixfigsizes]{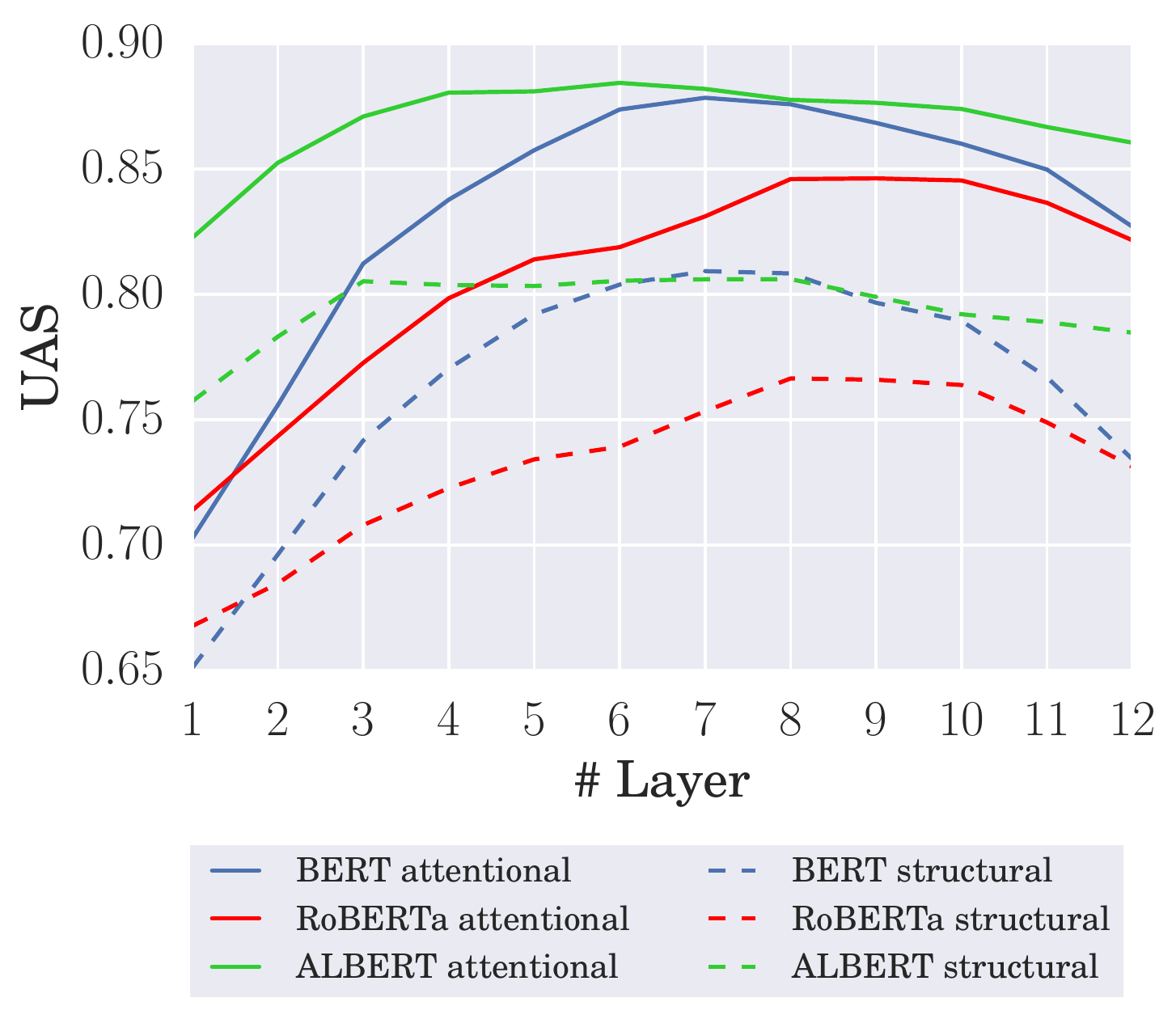}
     \end{subfigure}%
     \hfill
     \begin{subfigure}[b]{0.5\textwidth}
         \centering
         \includegraphics[width=\appendixfigsizes]{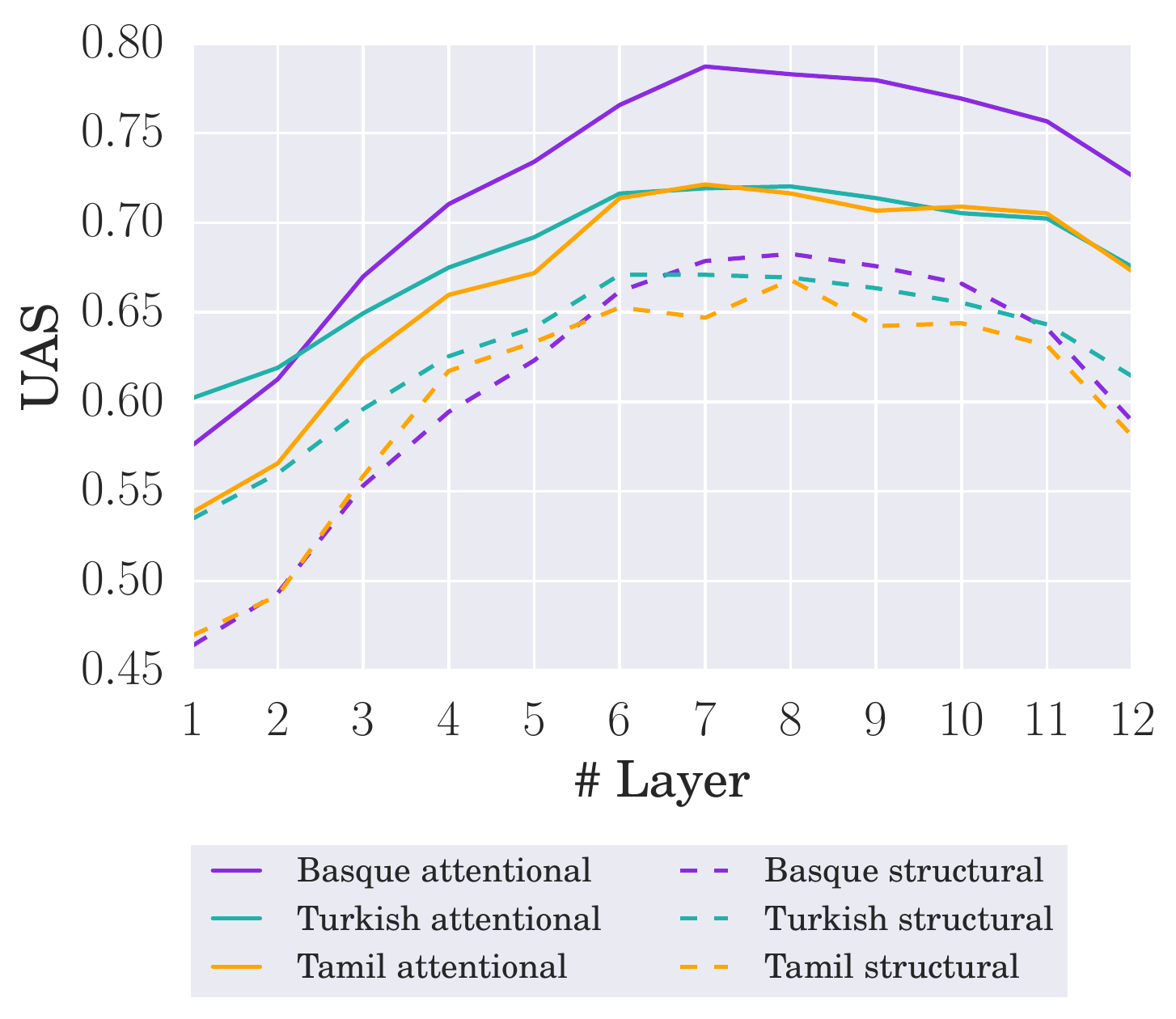}
     \end{subfigure}
    \caption{UAS of the probes evaluated on English using BERT, RoBERTa and ALBERT representations (left); Basque, Turkish and Tamil using BERT representations (right).}
    \label{fig:uas_results_appendix}
    \vspace{-10pt}
\end{figure}

\section{\binfo{} by Language}\label{app:vinfolang}

\begin{figure}[h]
    \centering
     \begin{subfigure}[b]{0.5\textwidth}
         \centering
         \includegraphics[width=\appendixfigsizes]{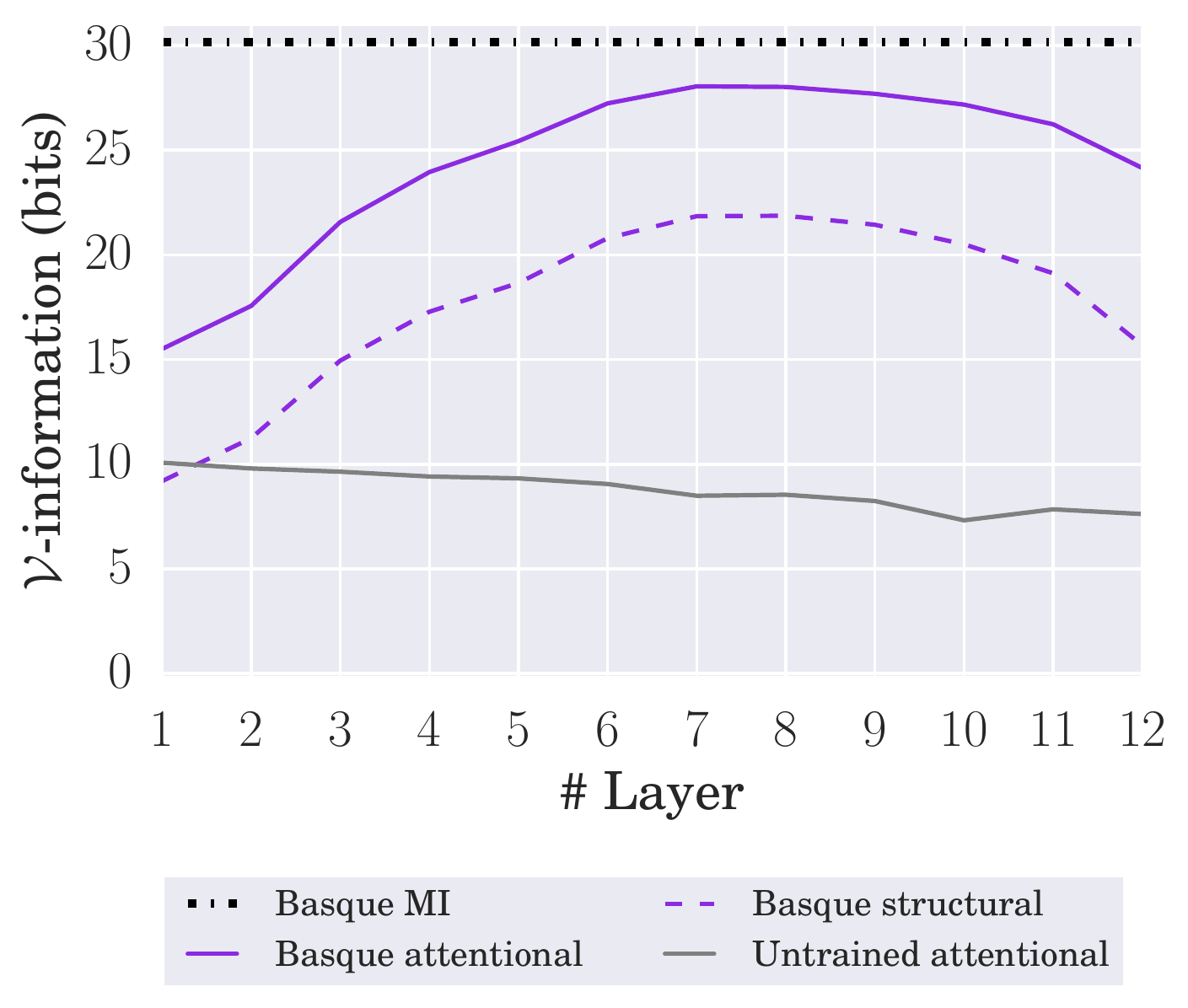}
     \end{subfigure}%
     \hfill
     \begin{subfigure}[b]{0.5\textwidth}
         \centering
         \includegraphics[width=\appendixfigsizes]{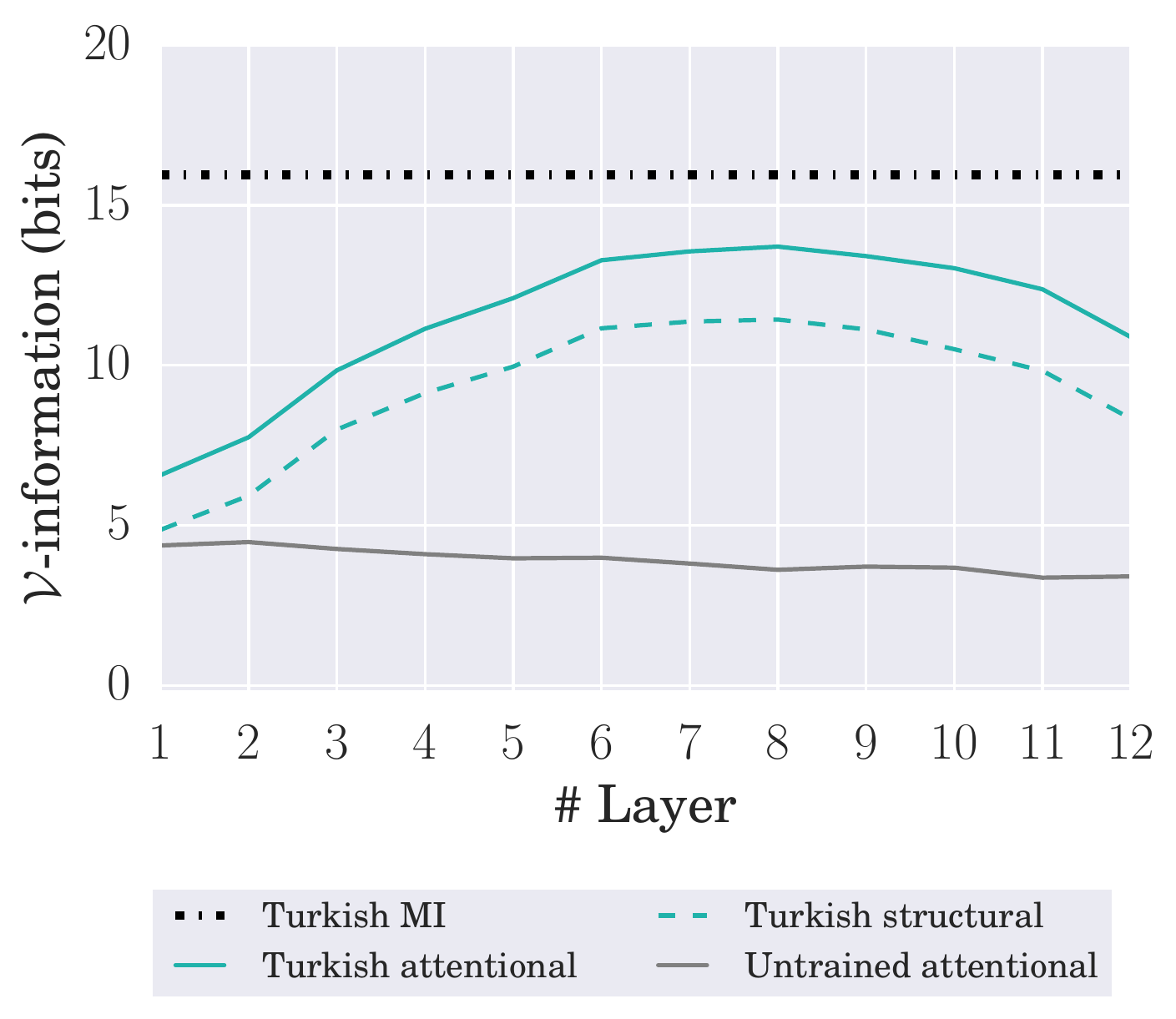}
     \end{subfigure}
     \begin{subfigure}[b]{0.5\textwidth}
         \centering
         \includegraphics[width=\appendixfigsizes]{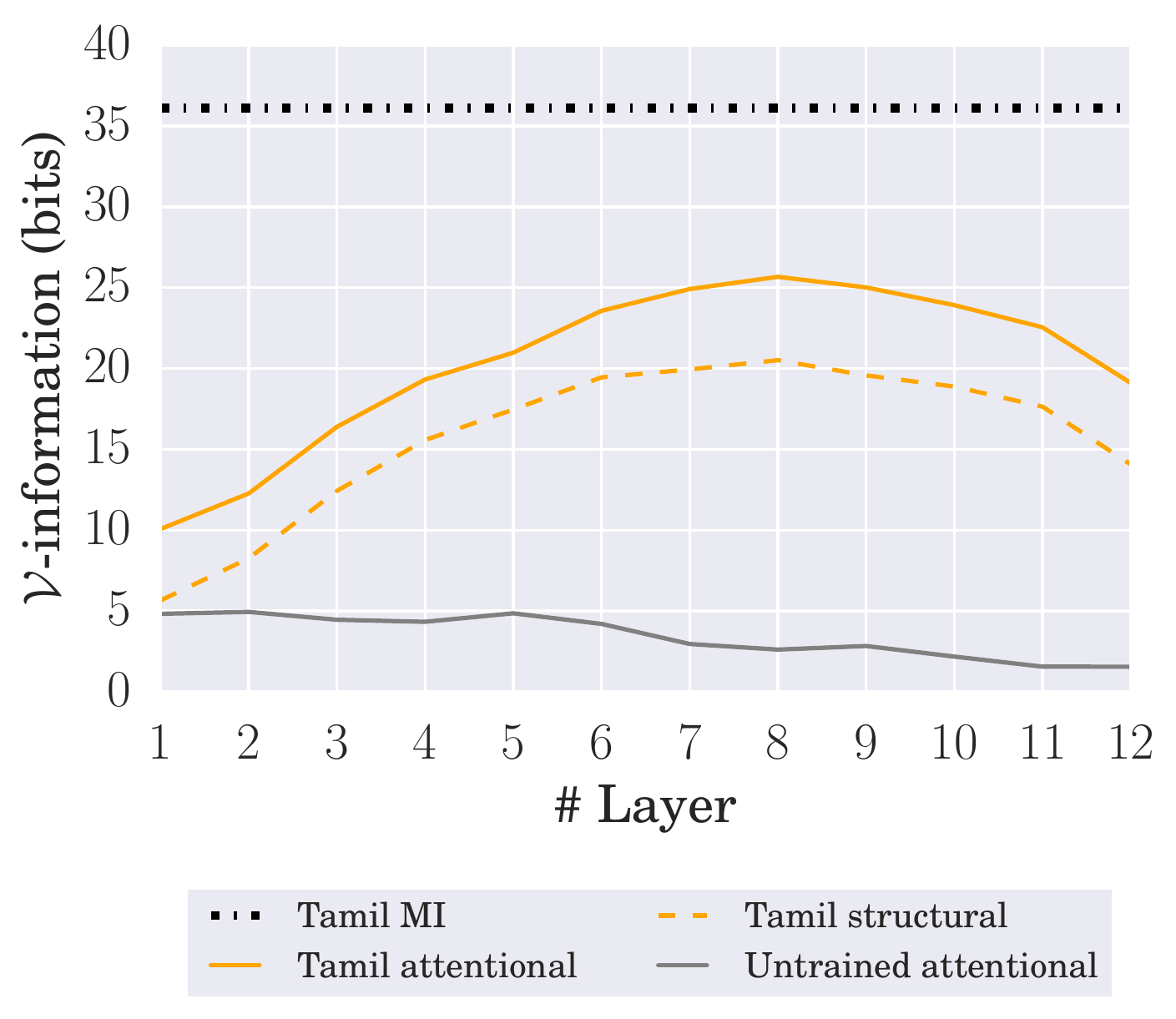}
     \end{subfigure}
    \caption{The estimated attentional \vinfo{}, structural \vinfo{}, and mutual information on Basque (top-left); Turkish (top-right); and Tamil (bottom).\looseness=-1}
    \label{fig:vinfo_all_langs}
    \vspace{-60pt}
\end{figure}

\clearpage

\section{Attention Head Weights Results}\label{app:logits}

\begin{figure}[h]
    \centering
     \begin{subfigure}[b]{0.5\textwidth}
         \centering
         \includegraphics[width=\appendixfigsizes]{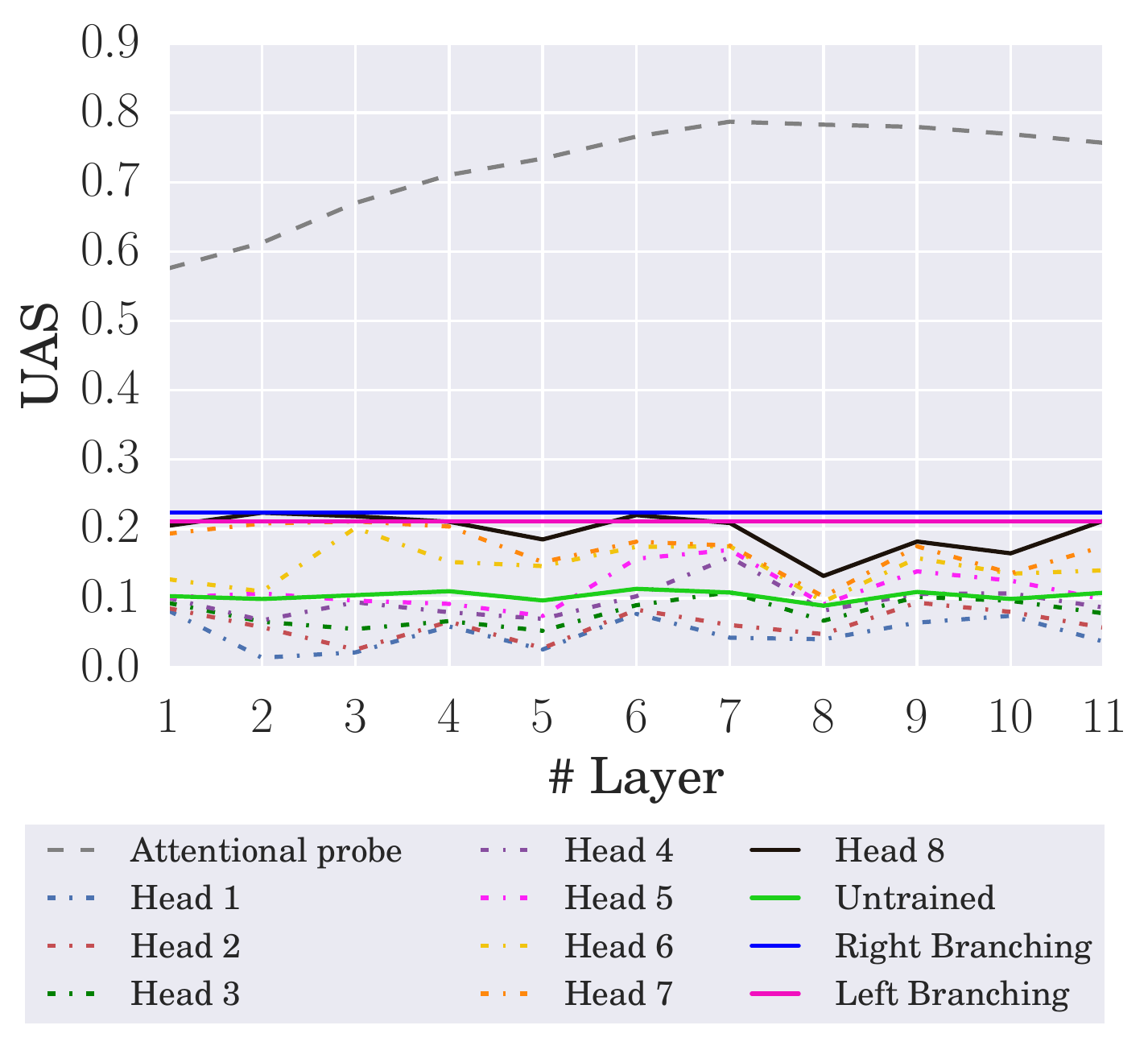}
     \end{subfigure}%
     \hfill
     \begin{subfigure}[b]{0.5\textwidth}
         \centering
         \includegraphics[width=\appendixfigsizes]{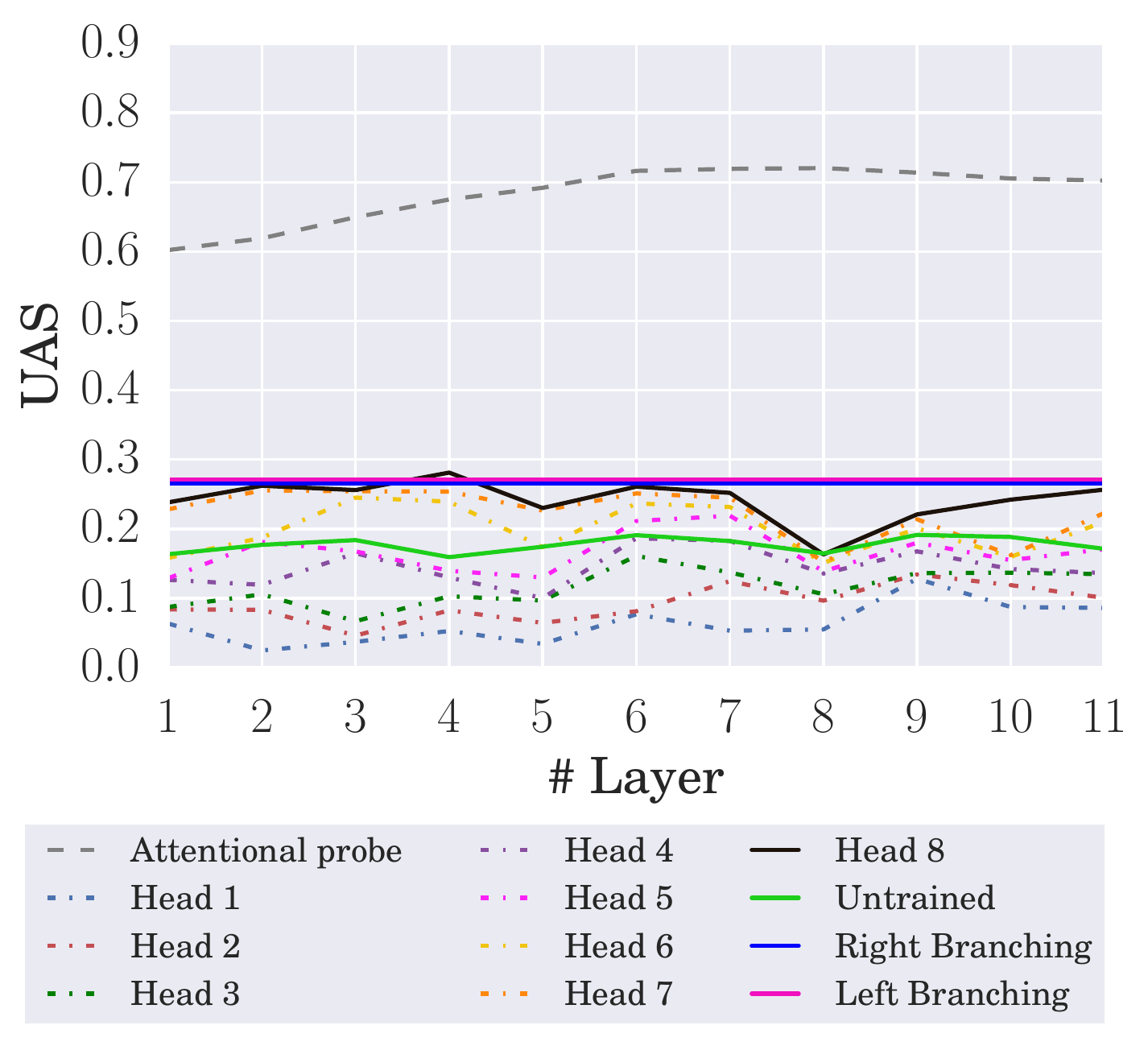}
     \end{subfigure}
     
     \begin{subfigure}[b]{0.5\textwidth}
         \centering
         \includegraphics[width=\appendixfigsizes]{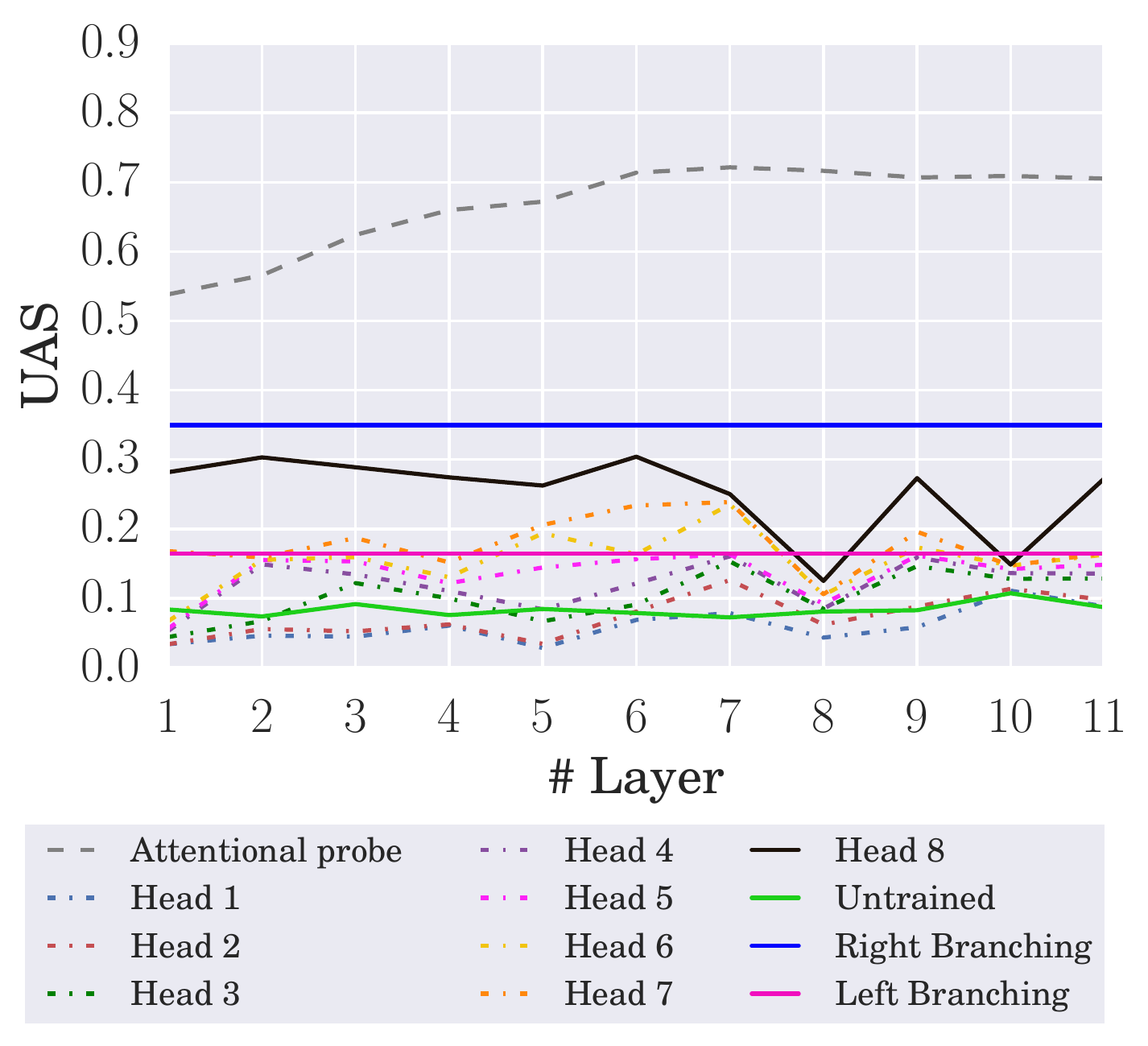}
     \end{subfigure}
     \caption{UAS of all attention heads' weights computed by BERT (with its pretrained parameters frozen) in Basque (top-left); Turkish (top-right); and Tamil (bottom).}
    \label{fig:uas_attention_heads}
\end{figure}

We additionally compute the parsing accuracy of the attention heads with their actual weights as taken from BERT (with its parameters as pretrained).\footnote{We assign the weight between each word and the root node as zero, since it is not part of the attention weights.}
In \Cref{fig:uas_attention_heads} (as well as \Cref{fig:english_head} in the main text),
we label the heads in order of their performance (1 is always the least accurate per layer, 8 the most). 
These results show that an attention head's potential to extract syntax trees is far above what each individual head actually extracts.

\section{Average Sentence Lengths}\label{app:len}
\begin{table}[h]
    \centering
    \begin{tabular}{lc}
        \toprule
        \textbf{Language} & \textbf{Average Sentence Length} \\
        \midrule
        Basque & 13 \\
        English & 15 \\
        Tamil & 17 \\
        Turkish & 10 \\
        \bottomrule
    \end{tabular}
    \caption{The average sentence length per language under consideration.}
    \label{tab:len}
\end{table}

\end{document}